\documentclass[sigconf]{acmart}

\usepackage{pifont}
\usepackage{latexsym}
\usepackage{balance}
\usepackage{amsmath}
\usepackage[ruled,linesnumbered]{algorithm2e} %
\usepackage{hyperref}
\usepackage{diagbox}
\usepackage{amsthm}
\usepackage{multirow}
\usepackage{pgffor}
\usepackage{epstopdf}
\usepackage{tikz}
\usepackage{epstopdf}
\usepackage{mathtools}
\usepackage{enumitem}
\usepackage{subfigure}
\usepackage{tabulary}
\usepackage{color}
\usepackage{flushend}
\usepackage{stfloats}
\usepackage{tcolorbox}
\usepackage[utf8]{inputenc}
\usepackage[english]{babel}
\usepackage{tabularx}
\usepackage{CJKutf8}
\usepackage{microtype}
\usepackage{textcomp}
\usepackage{booktabs}
\usepackage{colortbl}
\usepackage{soul}
\usepackage{appendix}
\usepackage[switch]{lineno}
\usepackage{xspace}
\usepackage{algpseudocode}
\usepackage{amsmath}

\newcommand{\mypara}[1]{{\noindent\textbf{#1}}}
\newcommand{\FedACK}{\textsc{FedACK}\xspace}
\DeclareMathOperator*{\argmin}{arg\,min}

\AtBeginDocument{%
  \providecommand\BibTeX{{%
    \normalfont B\kern-0.5em{\scshape i\kern-0.25em b}\kern-0.8em\TeX}}}

\copyrightyear{2023}
\acmYear{2023}
\setcopyright{acmlicensed}\acmConference[WWW '23]{Proceedings of the ACM
Web Conference 2023}{May 1--5, 2023}{Austin, TX, USA}
\acmBooktitle{Proceedings of the ACM Web Conference 2023 (WWW '23), May
1--5, 2023, Austin, TX, USA}
\acmPrice{15.00}
\acmDOI{10.1145/3543507.3583500}
\acmISBN{978-1-4503-9416-1/23/04}

\begin{document}

\title{\FedACK: Federated Adversarial Contrastive Knowledge Distillation for Cross-Lingual and Cross-Model Social Bot Detection}

\author{Yingguang Yang}
\affiliation{%
  \institution{USTC}
  \city{Heifei}
  \country{China}
}
\email{dao@mail.ustc.edu.cn}

\author{Renyu Yang}
\affiliation{%
  \institution{University of Leeds}
  \city{Yorkshire}
  \country{United Kingdom}
}
\email{r.yang1@leeds.ac.uk}
\authornote{Corresponding authors} 

\author{Hao Peng}
\affiliation{%
  \institution{Beihang University}
  \city{Beijing}
  \country{China}
}
\email{penghao@buaa.edu.cn}
\authornotemark[1]

\author{Yangyang Li}
\affiliation{%
  \institution{NERC-RPP, CAEIT}
  \city{Beijing}
  \country{China}
}
\email{liyangyang@cetc.com.cn}
\authornotemark[1]

\author{Tong Li}
\affiliation{%
  \institution{Tsinghua University}
  \city{Beijing}
  \country{China}
}
\email{tongli@mail.tsinghua.edu.cn}

\author{Yong Liao}
\affiliation{%
  \institution{USTC}
  \city{Heifei}
  \country{China}
}
\email{yliao@ustc.edu.cn}

\author{Pengyuan Zhou}
\affiliation{%
  \institution{USTC}
  \city{Heifei}
  \country{China}
}
\email{pyzhou@ustc.edu.cn}
\authornotemark[1]

\renewcommand{\shortauthors}{Y. Yang, R. Yang, H. Peng, et al.}

\begin{abstract}
Social bot detection is of paramount importance to the resilience and security of online social platforms. The state-of-the-art detection models are siloed and have largely overlooked a variety of data characteristics from multiple cross-lingual platforms. Meanwhile, the heterogeneity of data distribution and model architecture make it intricate to devise an efficient cross-platform and cross-model detection framework.
In this paper, we propose \FedACK, a new federated adversarial contrastive knowledge distillation framework for social bot detection. We devise a GAN-based federated knowledge distillation mechanism for efficiently transferring knowledge of data distribution  among clients. In particular, a global generator is used to extract the knowledge of global data distribution and distill it into each client’s local model. We leverage local discriminator to enable customized model design and use local generator for data enhancement with
hard-to-decide samples. Local training is conducted as multi-stage adversarial and contrastive learning to enable consistent feature spaces among clients and to constrain the optimization direction of local models,  reducing the divergences between local and global models. Experiments demonstrate that \FedACK outperforms the state-of-the-art approaches in terms of accuracy, communication efficiency, and feature space consistency.
\end{abstract}





\maketitle

\vspace{-0.6em}
\section{Introduction}

Social bots imitate human behaviors on social networks such as Twitter, Facebook, Instagram, etc.~\cite{yardi2010detecting}. Millions of bots, typically controlled by automated programs or platform APIs \cite{abokhodair2015dissecting}, attempt to sneak into genuine users as a disguise to pursue malicious goals such as actively engaging in election interference \cite{deb2019perils,ferrara2020characterizing}, misinformation dissemination \cite{DBLP:journals/cacm/Cresci20}, and privacy attacks \cite{varol2017online}. Bots are also involved in spreading extreme ideologies \cite{berger2015isis,ferrara2016predicting}, posing threats to online communities. Effective bot detection is necessitated by the jeopardized user experience on social media platforms and the induction of unfavorable social effects. 

There is a new yet understudied problem in bot detection -- a society of bots tend to be exposed to multiple social platforms and behave as collaborative cohorts. Existing bot detection solutions largely rely on user property features extracted from metadata~\cite{d2015real,yang2020scalable}, or features derived from textual data such as a tweet post \cite{wei2019twitter,feng2021satar}, before adopting graph-based techniques to explore neighborhood information \cite{zhao2020multi,DBLP:journals/corr/abs-2109-02927,yang2022rosgas}. While such models can uncover camouflage behaviors, they are siloed and subject to the amount, shape, and quality of platform-specific data. To this end, Federated Learning (FL) has becomes the main driving force of model training across heterogeneous platforms without disclosing local private datasets. Some studies \cite{zhu2021data,rasouli2020fedgan,zhang2022fine,zhang2022feddtg} augmented FL by Generative Adversarial Networks (GANs) and Knowledge Distillation (KD) in a data-free manner to safeguard privacy against intrusions. However, they have the following limitations:

i) \textit{Restriction to homogeneous model architecture.} As FL models assume homogeneous model architecture on a per client basis -- which however no longer holds -- participants are stringently required to conform to the same model architecture managed by a central server. It is therefore imperative to enable each individual platform to customize heterogeneous models as per unique data characteristics. ii) \textit{Inconsistent feature learning spaces}. The state-of-the-art Federated KD approaches are largely based on image samples and assume consistent feature space. However, the distinction between global and local data distribution tends to result in non-negligible model drift and inconsistent feature learning spaces, which will in turn cause performance loss. It is highly desirable to align feature spaces among different clients to improve the global model performance. 
iii) \textit{Sensitivity to content language}. Textual data based anomaly detection approaches to date are sensitive to the languages that the model is built upon. Existing solutions for cross-lingual content detection in online social networks either substantially raise computational costs \cite{du2021cross,zia2022improving,DBLP:journals/talip/DeBGE22} or require labor-intensive feature engineering to identify cross-lingual invariant features \cite{steimel2019investigating,dementieva2021cross,chu2021cross}. Arguably, how to incorporate into a synergetic model a variety of customized models with heterogeneous data in different languages to enable consistent feature learning space is still under-explored.  



\begin{figure}[t]
\centering
\includegraphics[width=0.34\textwidth]{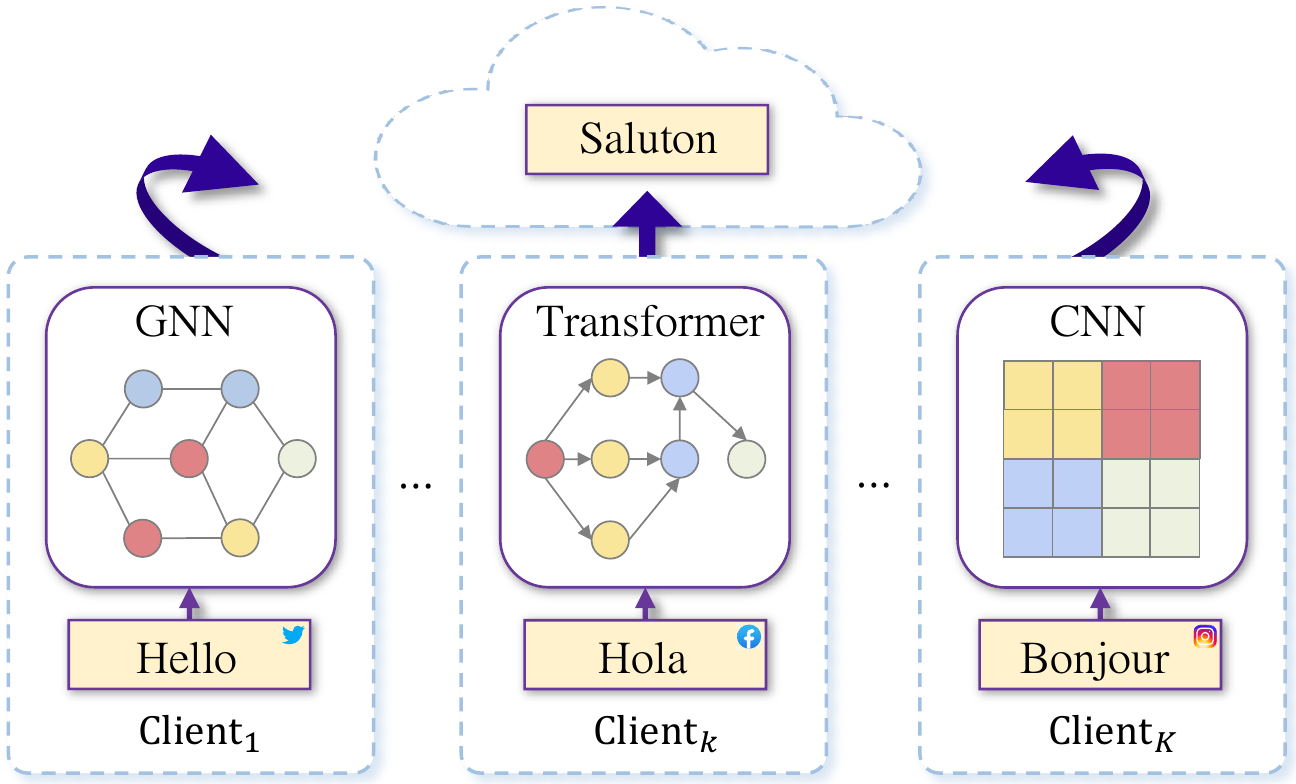}
\vspace{-0.8em}
\caption{Incorporating multiple social platforms with heterogeneous languages, context spaces and model architectures}
\vspace{-0.5em}
\label{fig:motivation} 
\end{figure}

This paper proposes \FedACK, a novel bot detection framework through \underline{Fed}erated \underline{A}dversarial learning \underline{C}ontrastive learning and \underline{K}nowledge distillation.  \FedACK envisions to enable personalization of local models in a consistent feature space across different languages (see Fig.~\ref{fig:motivation}). We present a new federated GAN-based knowledge distillation architecture -- a global generator is used to extract the knowledge of global data distribution and to distill the knowledge into each client's local model. We elaborate two discriminators -- both globally shared and local -- to enable customized model design and use a local generator for data enhancement with hard-to-decide samples. 
Specifically, the local training on each client side is regarded as a multi-stage adversarial learning procedure to efficiently transfer data distribution knowledge to each client and to learn consistent feature spaces and decision boundaries. We further exploit contrastive learning to constrain the optimization direction of local models and reduce the divergences between local and global models.  To replicate non-IID data distribution across multi-platforms, we employ two real-world Twitter datasets, partitioned by the Dirichlet distribution. Experiment shows that \FedACK outperforms the state-of-the-art approaches on accuracy and achieves competitive communication efficiency and consistent feature space. This work makes the following contributions.

\begin{itemize}[leftmargin=*]
    \item To the best of our knowledge, \FedACK is the first social bot detection solution based on federated knowledge distillation that envisions cross-lingual and cross-model bot detection.
    \item contrast and adversarial learning mechanisms for enabling consistent feature space for better knowledge transfer and representation when tackling non-IID data and data scarcity among clients.
    \item \FedACK outperforms other FL-based approaches by up to 15.19\% accuracy improvement in high heterogeneity scenarios, and achieves up to 4.5x convergence acceleration against the 2nd fastest method.
\end{itemize}
 
To enable replication and foster research, \FedACK is publicly available at: \texttt{https://github.com/846468230/FedACK}.

\vspace{-0.6em}
\section{Preliminaries}
\vspace{-0.1em}

\subsection{Background}
\mypara{Federated learning (FL).} FL 
is a distributed learning paradigm that allows clients to perform local training before aggregation without sharing clients' private data \cite{DBLP:journals/corr/KonecnyMRR16,mcmahan2017communication,bonawitz2017practical,peng2021differentially,liu2022federated}.
While promising, FL can have inferior performance particularly when training data is not independent and identically distributed (Non-IID) on  local devices~\cite{zhao2018federated,li2020federated}, which could make the model deflected to a local optimum \cite{karimireddy2020scaffold}.
Most of the existing works mainly fall into two categories. The first is introducing additional data or using data enhancement to address model drift issue caused by the non-IID data. 
FedGAN \cite{rasouli2020fedgan} trains a GAN to tackle the non-IID data challenge in a communication efficient way but inevitably produces bias.
FedGen \cite{zhu2021data} and FedDTG \cite{zhang2022feddtg} utilize generator to simulate the global data distribution to improve performance.
The second category mainly focuses on local regularization.
FedProx \cite{li2020federated} adds an optimization item to local training, and SCAFFOLD \cite{karimireddy2020scaffold} uses control variants to correct the \textit{client-drift} in local updates while guaranteeing a faster convergence rate.
FedDyn \cite{DBLP:conf/iclr/AcarZNMWS21} and MOON \cite{li2021model} constrain the direction of local model updates by comparing the similarity between model representations to align the local and global optimization objectives. However, these approaches either direct model aggregation to get the global model that causes non-negligible performance deterioration \cite{singh2020model} or neglect the impact of data heterogeneity, which may lead to knowledge loss of local data distribution during the model aggregation.

\mypara{Federated knowledge distillation (KD)}. KD is first introduced to use compact models to approximate the function learned by larger models \cite{bucilu2006model}. Knowledge is formally referred to as the softened logits, and in a typical KD, a student model absorbs and mimics the knowledge from teacher models \cite{hinton2015distilling}. KD is inherently beneficial for FL since it requires less or no data to enable the model to understand the data distribution. FedDistill \cite{seo2020federated} jointly refines logits of user-data obtained through model forward propagation and performs global knowledge distillation to reduce the global model drift problem.  FedDF \cite{lin2020ensemble} proposes an ensemble distillation for model fusion and trains the global model through averaged logits from local models.
FedGen \cite{zhu2021data} combines each local model's average logits as the teacher in KD to train a global generator. FedFTG \cite{zhang2022fine} uses each local model's logit as the teacher to train a global generator and distill knowledge by using the pseudo-data generated by the global generator to fine tune the global model. However, none of them focuses on enabling consistent feature space, which will lead to ineffective knowledge dissemination. FL and KD have been largely overlooked to date in social bot detection, which is investigated in a siloed way~\cite{DBLP:journals/cacm/Cresci20}. \FedACK can fill this gap through enhanced adversarial learning with a shared discriminator and an exclusive discriminator to support designated cross-model bot detection. 

\mypara{Cross-lingual content detection in social networks}.
Publishing fake or misleading contents through social bots on social networks in different languages has become the norm rather than the exception. \cite{dementieva2021cross,chu2021cross} have explored the possibilities of cross-lingual content detection by seeking cross-lingual invariant features. There is also a huge body of research on cross-lingual text embedding and model representation \cite{DBLP:conf/acl/Nozza20,du2021cross,zia2022improving,DBLP:journals/talip/DeBGE22,peng2022reinforced} for detecting hate speeches, fake news or abnormal events. These works usually require huge efforts in finding cross-lingual invariants in the data, and thus computational inefficiency. While InfoXLM \cite{chi2020infoxlm} could be applied in \FedACK as a substitute for our cross-lingual module, it may involve additional overhead given only a few mainstream languages in social platforms. \FedACK implemented text embedding by mapping the cross-lingual texts into the same context space.

\vspace{-0.48em}
\subsection{Problem Scope}

We consider federated social bot detection setting that includes a central server and \textit{K} clients holding private datasets $\{\mathcal{D}_1,\dots,\mathcal{D}_K\}$. These private datasets contain benign accounts and different generations of bots. 
Presumably, different model architectures or parameters exist on different clients. \FedACK focus on meta and text data, rather than multimodal data. Instead of collecting raw client data, the server tackles heterogeneous data distribution across clients and aggregates model parameters for the shared networks. 
The objective is to minimize the overall error among all clients:
\begin{equation}
\vspace{-0.2em}
    \argmin \limits_{w} \mathcal{L}(w)= \frac{1}{K} \sum\limits^K_{k=1}  \frac{1}{N_k}\sum\limits^{N_k}_{i=1}\mathcal{L}(x^k_i,y^k_i;w), \label{lossCE}
\vspace{-0.2em}
\end{equation}
where $\mathcal{L}$ is the loss function that evaluates the prediction model $w$ on the data sample $(x^k_i,y^k_i)$ of $\mathcal{D}_k=\{(x^k_i,y^k_i)\}{ |}^{N_k}_{i=1}$ in $k$-th client.

\vspace{-0.3em}
\section{Methodology}
\begin{figure*}[t]
\centering
\includegraphics[width=0.9\textwidth]{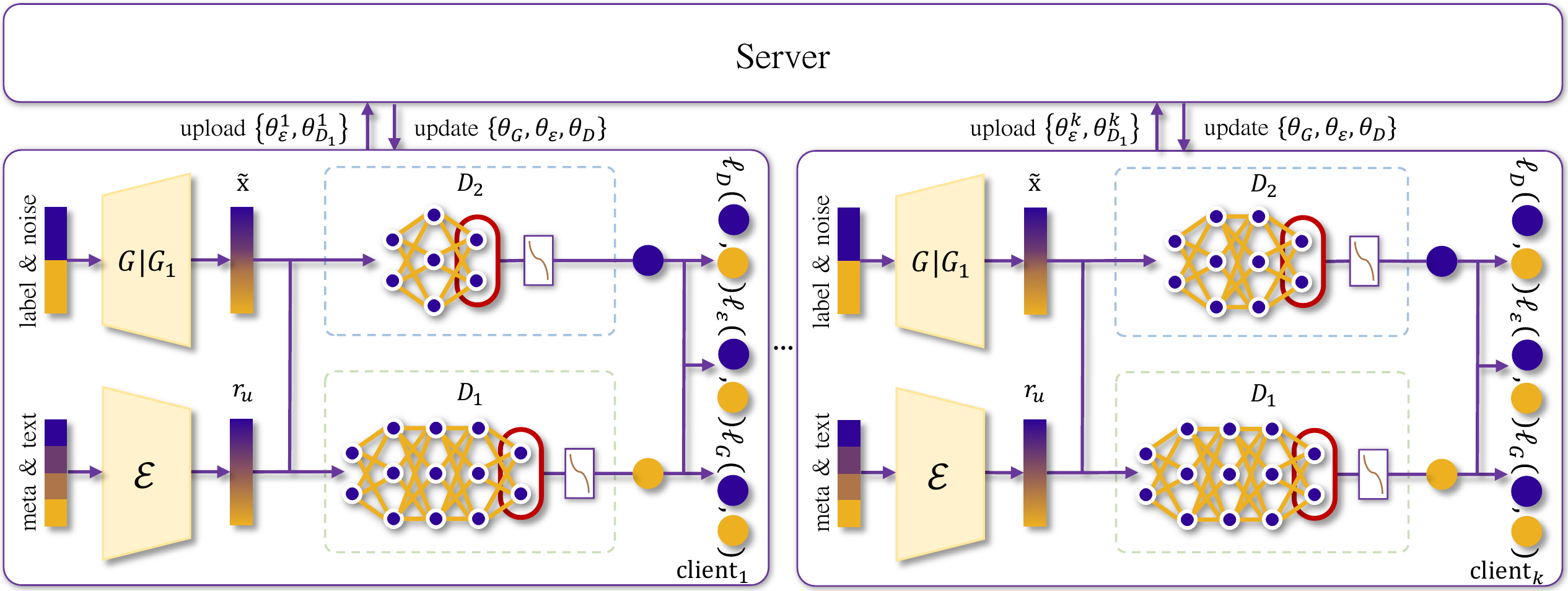}
\vspace{-0.8em}
\caption{The proposed \FedACK framework.}
\label{fig:framework}
\end{figure*}

As shown in Fig.~\ref{fig:framework}, \FedACK consists of cross-lingual mapping, backbone model, and federated adversarial contrastive KD.

\vspace{-0.4em}
\subsection{Cross-Lingual Mapping}

We adopt a Transformer encoder-decoder-based method to achieve the alignment of different language contents. 
In essence, given a $m$-word text $x=\{x_1,...,x_m\}$ in one language and the corresponding $n$-word text $y=\{y_1,...,y_n\}$ in another language, we use an \textbf{Encoder $\phi_E$} to transform the source text $x$ and the target text $y$ into context representations $z_x=\{z_{x_1},...,z_{x_m}\}$ and $z_y=\{z_{y_1},...,z_{y_n}\}$. We devise a \textbf{Mapper $\mathcal{M}$} for converting between two context representation spaces, i.e., ${z_x}'=\mathcal{M}(z_y)$ and ${z_y}'=\mathcal{M}(z_x)$. 

We introduce an adversarial mechanism for optimizing $\mathcal{M}$ so that the original $z_x$ and the mapped ${z_x}'$ can be sufficiently similar. We first obtain the embedding of the context representations:
\begin{equation}
    \bar{z}_x=\frac{1}{m}\sum\limits_{k=1}^m{z_{x_k}},{\bar{z}_x}'=\frac{1}{n}\sum\limits_{k=1}^n{z_{x_k}'}. \label{eqmeanrep}
\end{equation}
Then we use the discriminator $D$ to distinguish whether an embedding $\bar{z}_x$ or ${\bar{z}_x}'$ is forward propagated from $\mathcal{M}$ ($\bar{z}_y$ is equal to ${\bar{z}_y}'$). Accordingly, the loss of the discriminator is defined as:
\begin{equation}
\begin{aligned}
    &\mathcal{L}_{dis} = \mathcal{L}_{dis_x} + \mathcal{L}_{dis_y}, \\
    \mathcal{L}_{dis_x} = & {(D(\bar{z}_x)-y_{\bar{z}_x})^2+(D({\bar{z}_x}')-y_{{\bar{z}_x}'})^2} \label{disx},
\end{aligned}
\end{equation}
where the label $y_{\bar{z}}$ is set as 0 if an embedding $\bar{z}$ is mapped from $\mathcal{M}$. 
We combine the encoder $\phi_E$ and the mapper $\mathcal{M}$ into the generator. Similarly, the loss of the generator is:
\begin{equation}
    \mathcal{L}_{gen} = (D({\bar{z}_x}')-y_{{\bar{z}_x}'})^2 +(D({\bar{z}_y}')-y_{{\bar{z}_y}'})^2, \label{genx}
\end{equation}
where the label $y_{\bar{z}}$ is always set as 1 to produce sufficiently similar representations to confuse the discriminator.
We collect the context representations $(z_x,z_y,{z_x}',{z_y}')$ and decode them with a Decoder $\phi_D$ and generate the corresponding translation $(\tilde{z}_y,\tilde{z}_x,{\tilde{z}_y}',{\tilde{z}_x}')$, where $\tilde{z}_y=\phi_D(z_x)$ and other terms are calculated in a similar way.

The loss of the Transformer is therefore defined as the cross-entropy loss between $x$ and $y$:
\begin{equation}
\begin{aligned}   
    \mathcal{L}_{trans} = \mathcal{L}_{z_x} + \mathcal{L}_{z_y} + \mathcal{L}_{{z_x}'} + \mathcal{L}_{{z_y}'}, \\
    \mathcal{L}_{z_x} = -\sum\limits_{t=1}^n\log P(y_t|\tilde{z}_y<t,z_x), \\
    \mathcal{L}_{{z_x}'} = -\sum\limits_{t=1}^n\log P(y_t|{\tilde{z}_y}'<t,{z_x}'). \label{crosslingualloss}
\end{aligned}
\end{equation}

\subsection{Backbone Model for Feature Extraction}
The main task is to let user-level backbone feature extractor model $\varepsilon$ to extract features per user metadata (e.g., account properties) and textual data (e.g., tweets).
We concatenate the key items extracted from the metadata into the property vector $u_p$, following the similar way as \cite{yang2020scalable,yang2022rosgas}, which is converted into user's property representation $r_p$ by a Multi-layer Perceptron $r_p = MLP(u_p)$. For textual data, assume a user has posted M Tweets $u_t=\{t_1^1,\dots,t_{Q_1}^1,t_1^2,\dots,t_{Q_M}^M\}$, possibly in the form of different languages; $t^j_i$ represents the $i$-th word in the $j$-th tweet. We leverage
the Encoder and Mapper delivered by the aforementioned cross-lingual module to convert each $u_t$ into a uniformed contextual space, i.e.,  $z_{u_t}=\{z_1^1,\dots,z_{Q_1}^1,{z_1^2}',\dots,z_{Q_M}^M\}$. A Convolutional Neural Networks (CNN) layer, i.e., TextCNN\cite{DBLP:conf/emnlp/Kim14}, is then used to obtain the tweet-level representation $h_{u_t}$:
\begin{equation}
    h_{u_t} = \{h_1^t,\dots,h_M^t\},h_j^t = CNN(\{z_1^j,\dots,z_{Q_1}^j\}).
\end{equation}
An attention layer is used to quantify the influence of each tweet on the overall semantics of the user 
and to calculate the user-level tweet representation $r_t$ through weighted aggregation of all tweets.
The complete user-level representation $r_u$ is:
\begin{equation}
    r_u = MLP(concat(r_p,r_t)).
\end{equation}

\subsection{Federated Adversarial Contrastive KD}

Architecturally, we follow the conventional server-client design for the Federated GAN-based KD. A client $k$ contains a global generator $G$ for KD, and a local generator $G_k$ for data enhancement. 
We use two discriminators -- $D_1$ shared  across all clients with the same architecture and initial parameters but trained on local data, and $D_2$ exclusively designated for each client to satisfy its individual demand.
Each client uses a backbone model $\varepsilon$ to get the user's representation $r_u=\varepsilon (x_u)$ from local data $\mathcal{D}_k$. As a significant departure from the state-of-the-arts, a new multi-stage adversarial mechanism is proposed for jointly optimizing classification in both discriminators at intra-client level and a contrastive mechanism for aligning different feature spaces across clients. The notations are detailed in Appendix \ref{sec:append:notation}.

\vspace{-0.5em}
\subsubsection{Local Adversarial Contrastive KD}
The adversarial learning on a per client basis include the following multi-phases.



\mypara{Stage-1: Training D1 and D2 as classifiers.} We aim to facilitate the two  models to learn different decision boundaries for the same class of samples and compress the feature space of the feature extractor. To tackle \textit{non-IID} data distribution and data \textit{scarcity} among clients, we treat the shared global generator $G$ as teacher network and distill the knowledge of global data distribution from it. For each sample $(x^k_i,y^k_i)$, $G$ uses a standard Gaussian noise $z\sim\mathcal{N}(0,1)$ and label $y^k_i$ to generate pseudo-data $\tilde{x}=G(z,y^k_i;\theta_G)$. 
$(x^k_i,\tilde{x})$ is fed into $D_1$ to obtain the probability distributions $(p,\tilde{p})$.
We make $D_1$ fit probability $p$ of real data $x^k_i$ close to $\tilde{p}$ to distill knowledge by minimizing Eq.~(\ref{distill}):
\begin{equation}
\vspace{-0.4em}
 \mathcal{L}^k_{dis} = \frac{1}{N_k}\sum\limits^{N_k}_{i=1}
    D_{KL}(\sigma(D_1(r^k_i))||\sigma(D_1(\tilde{x})) \label{distill},
\vspace{-0.2em}
\end{equation} 
where $\sigma$ is the softmax function, $D_{KL}$ is the Kullback–Leibler divergence and $r^k_i= \varepsilon(x^k_i)$. Similar ${\mathcal{L}^k_{dis}}'$ is performed for $D_2$. 

We induce an \textit{adversarial loss} to measure the probability difference between $D_1$ and $D_2$: 
\begin{equation}
    \mathcal{L}^k_{adv} = \frac{1}{N_k}\sum\limits^{N_k}_{i=1} D_{KL}(\sigma(D_1(r^k_i))||\sigma(D_2(r^k_i))). \label{lossadv}
\end{equation}
In essence, maximizing Eq. (\ref{lossadv}) can produce distinct decision boundaries between two discriminators to compress overlapping decision space. Intuitively, this is because $\varepsilon$ would learn more precise feature space if the representations could be correctly classified by both discriminators. Similarly to Eq.~(\ref{lossadv}), we calculate adversarial loss $\mathcal{L}^k_{advg}$ on the pseudo-data generated by the local generator $G_k$ and minimize it to intensify the above process. 
$D_1$ and $D_2$ also need to correctly classify the pseudo-data randomly generated by $G$ to deal with the issue of unbalanced and scarce data. 
To sum up, the overall loss function of the discriminators is:
\begin{equation}
    \mathcal{L}^k_D = \mathcal{L}^k_{cls} + \alpha(\mathcal{L}^k_{dis}+{\mathcal{L}^k_{dis}}') +  \gamma(\mathcal{L}^k_{advg} -\mathcal{L}^k_{adv}). \label{lossD}
\end{equation} 

\mypara{Stage-2: Training $\varepsilon$.} We minimize the adversarial loss Eq.~(\ref{lossadv}) to reduce the difference between the probability outputs of two discriminators for the same feature. This means that the same data point can fall on the same side of the decision boundaries of the two discriminators. The feature space of the feature extractor is compressed and enforced to generate more precise features. 

We use contrastive learning to navigate the optimization direction of feature extractor, thereby overcoming the \textit{model drift} between local extractors and the global extractor. We expect the newly optimized $\varepsilon^k_{t}$ to produce a representation $r=\varepsilon^k_{t}(x^k_i)$ as close as possible to the $r_{glo}=\varepsilon_{t}(x^k_i)$ generated by the global extractor $\varepsilon$ while as far away as possible from the last round result of the feature extractor $r_{pre}=\varepsilon_{t-1}(x^k_i)$.
We define the following contrastive loss function:
\begin{equation}
    \mathcal{L}^{k,i}_{con} = -\log \frac{\exp({\rm sim}(r,r_{glo})/\tau)}
    {\exp({\rm sim}(r,r_{glo})/\tau)+\exp({\rm sim}(r,r_{pre})/\tau)}, \label{losscon}
\end{equation} 
where \textit{sim} is the similarity measure function and $\tau$ denotes a temperature parameter. This can not only reduce the local models' drift but also serve as a bridge with adversarial learning to make models of different clients have a consistent feature space. The loss function extractor is defined as follows:
\begin{equation}
\vspace{-0.2em}
    \mathcal{L}^k_{\varepsilon} = \mathcal{L}^k_{cls} +   \gamma\mathcal{L}^k_{adv}+ \mu \frac{1}{N_k}\sum\limits^{N_k}_{i=1}\mathcal{L}^{k,i}_{con}. 
    \vspace{-0.2em}
\label{losse}
\end{equation}


\mypara{Stage-3: Training $G_k$}. We maximize $\mathcal{L}^k_{advg}$ to ensure $G_k$ can generate pseudo-data that falls near the decision boundaries of two discriminators. This enforces such boundaries closer to the coincidence region, which further compresses the consistent feature space of the $\varepsilon$. As $G_k$ only generates the same data for each class, we add the diversity loss $\mathcal{L}_{var}$ to improve the diversity of the generated data and prevent model collapse:
\begin{equation}
\vspace{-0.4em}
\begin{aligned}
    \mathcal{L}_{var} = e^{\frac{1}{N*N}\sum\limits_{i,j\in\{1,\cdots,N\}}(-\parallel\tilde{x}_i-\tilde{x}_j\parallel_2*\parallel z_i-z_j\parallel_2)},
\end{aligned}   
\end{equation}
where $\tilde{x}_i=G_k(z_i,\hat{y}_i)$. The overall loss for generator $G_k$ can be calculated through:
\begin{equation}
    \mathcal{L}^k_{g} = \mathcal{L}^k_{cls}- \mathcal{L}^k_{advg} + \mathcal{L}_{var}.  \label{lossgen}
\end{equation}

\subsubsection{Server Aggregation Knowledge Extraction} In each communication round, the client $k$ uploads the local parameters of $\{\theta^k_\varepsilon,\theta^k_{D_1}\}$ to the server once the local training is finished, then waits for the updated global $G$, $D$ and $\varepsilon$ from the server to start a new round of local training. Once the server receives the latest parameters of the participating clients, it performs the model aggregation by weighted average and gets the updated global $\varepsilon$ and $D$:
\begin{equation}
    \theta_\varepsilon = \sum^M_{k=1} \frac{\parallel\mathcal{D}_k\parallel}{\parallel\mathcal{D}\parallel} \theta^k_{\varepsilon}, 
    \theta_D = \sum^M_{k=1} \frac{\parallel\mathcal{D}_k\parallel}{\parallel\mathcal{D}\parallel} \theta^k_{D_1},
    \label{aggregateEpsilon}
\end{equation}
where $M$ is the number of participating clients. 

To ensure global data distribution can adapt to local distributions that may largely drift from each other, we exploit the global discriminator $D$ and global generator $G$ for global knowledge extraction, without a need for server-side KD using proxy data. client's $D_1$ is used as the teacher network and define the loss of $G$ as:
\begin{equation}
\vspace{-0.4em}
\begin{aligned}
    \mathcal{L}_{G} =\frac{1}{K}\sum^K_{k=1} \sum_{\tilde{x}} \alpha^{k,y}_t[D_{KL} \left( \sigma(D^k_1(\tilde{x}))||\sigma(D(\tilde{x})) \right)  + \mathcal{L}^{\tilde{x}}_{cls}], \label{lossG}
    \end{aligned}
\end{equation}
where $\tilde{x}$ is from empirical samples $\mathcal{D}_G$ generated by $G$ using noise $z\sim\mathcal{N}(0,1)$ and label $y\sim p(y)$. $\alpha^{k,y}_t$ is the ratio of samples with label $y$ stored in client $k$ against the same label samples in $\mathcal{D}$. $p(y)$ is obtained by label counts from clients through communication.  

\vspace{-0.5em}
\subsection{Pipeline of \FedACK}
Alg.~\ref{algorithm1} summarizes the overall pipeline of \FedACK. The cross-lingual model $\phi_E$ and $\mathcal{M}$ for the backbone model $\varepsilon$ are first trained on the server (Line \ref{trainepsilon}) and distributed to all clients.
In each communication round, \FedACK first broadcasts the up-to-date $G$, $\varepsilon$ and $D$ to a selected subset of clients $S_t$ (Line \ref{algl:broadcast}). Each client optimizes all the required models $D^k_1, D^k_2$, $\varepsilon$ and $G_k$ using local data (Lines \ref{trainD}-\ref{trainGk}). When the parallel optimization completes, the server aggregates the clients' parameters in this round to update the global parameters $\theta_\varepsilon,\theta_D$ and to optimize the global generator $G$ (Lines \ref{Alg:aggragetepslon}-\ref{trainG}).

\begin{algorithm}[t]
\small
\caption{\FedACK}\label{algorithm1}
\KwIn{
Local data $\mathcal{D}_k$, $k=1,\cdots,K$, corpus data $\mathcal{D}_c$, models $G_k$, $\phi_E$, $\mathcal{M}$, $\varepsilon$, $D^k_1$, $D^k_2$, $G$, $D$
}
\KwOut{Local model $D_2$, global models $\phi_E$, $\mathcal{M}$, $\varepsilon$, $G$ }
initialization\;
Server trains $\phi_E$, $\mathcal{M}$ via Eq.(\ref{crosslingualloss}) based on $\mathcal{D}_c$, and transfers $\theta_{\phi_E}$, $\theta_{\mathcal{M}}$ to clients\; \label{trainepsilon}
\For{\rm {each communication round} $t=1,\ldots,T$}
{
    
     $S_t \gets$ random subset ($C$ fraction) of the $K$ clients\; \label{algl:6}
     Server broadcasts $\{\theta_G, \theta_\varepsilon, \theta_{D}\}$ to $S_t$ \;\label{algl:broadcast}
     \For{\rm {each client} $k\in S_t$ \rm{\textbf{in parallel}}}{
        Client $k$ updates $\{\theta_G, \theta^k_\varepsilon, \theta^k_{D_1}\}$ \;
        \For{\rm{each local epoch} $e=1,\cdots,E$ \label{trainD}
        }{ calculate $\mathcal{L}^k_D$ via Eq.(\ref{lossD}) \;
            \{$\theta^k_{D_1},\theta^k_{D_2}\} \gets \{\theta^k_{D_1},\theta^k_{D_2}\}-\nabla\mathcal{L}^k_D$\;
        }
        \For{\rm{each local epoch} $e=1,\cdots,E$ \label{trainE}
        }{ calculate $\mathcal{L}^k_{\varepsilon}$ via Eq.(\ref{losse}),
            $\theta^k_\varepsilon \gets \theta^k_\varepsilon -\nabla\mathcal{L}^k_{\varepsilon}$\;
        }
        \For{\rm{each local epoch} $e=1,\cdots,E$ \label{trainG}
        }{ calculate $\mathcal{L}^k_{g}$ via Eq.(\ref{lossgen}), $\theta^k_{G_k} \gets \theta^k_{G_k}-\nabla\mathcal{L}^k_{g}$\; 
        }\label{trainGk}
        Client $k$ sends $\{\theta^k_\varepsilon,\theta^k_{D_1}\}$ back to server\;
     }
    Server update $\{\theta_\varepsilon,\theta_D\}\gets$  Eq.(\ref{aggregateEpsilon}) \label{Alg:aggragetepslon} \;
    Server calculate $\mathcal{L}_{G}$ for $G$ via Eq.(\ref{lossG}) \label{trainG} \;
    $\theta_G\gets\theta_G-\nabla\mathcal{L}_{G}$
} \label{algl:24}
\end{algorithm}
\vspace{-0.3em}
\section{Evaluation}

The experiments aim to answer the following questions:
\begin{itemize}[leftmargin=*]
    \item \textbf{Q1}. How does \FedACK perform in classification  under different data distribution scenarios?
    \item \textbf{Q2}. How does \FedACK perform in learning efficiency?
    \item \textbf{Q3}. Can \FedACK  learn consistent feature space across clients?
    \item \textbf{Q4}. What is the effect of the different parameter values in different stages of \FedACK?
    \item \textbf{Q5}. How does the cross-lingual module perform when combined \FedACK and other baselines?
\end{itemize}

\vspace{-0.7em}
\subsection{Experimental Setup}
\label{exp:setup}

\subsubsection{Software and Hardware}
\FedACK is implemented with Python 3.8.10, Pytorch 1.7.1 and runs on two servers, one is equipped with NVIDIA Tesla V100 GPU, 2.20GHz Intel Xeon Gold 5220 CPU and 512GB RAM, and the other is equipped with NVIDIA GeForce RTX 3090 GPU, 3.40GHz Intel Xeon Gold 6246 CPU and 256GB RAM.

\vspace{-0.2em}
\subsubsection{Datasets} 
We conduct experiments on two Twitter bot datasets \textbf{Vendor-19} \cite{yang2019arming} and \textbf{TwiBot-20} \cite{feng2021twibot}, the largest ones in the public domain by far. 
We mix the \textbf{Vendor-19} with a dataset of benign accounts \textbf{Verified} which is presented in \cite{yang2020scalable}.
The newly released \textbf{TwiBot-20} dataset exposes users' social relationships and enables the use of advanced graph representation-based algorithms. More dataset statistics are outlined in Appendix \ref{sec:append:datasets}.

\vspace{-0.2em}
\subsubsection{Data heterogeneity} We use Dirichlet Distribution Dir ($\alpha$) to mock the non-IID given in \cite{li2021federated} and split the bot dataset with heterogeneity. $\alpha$ is an indicator of Dirichlet distribution -- the smaller $\alpha$ is, the more heterogeneous the data distribution is.

\vspace{-0.2em}
\subsubsection{Baselines} 
As federated KD provided a natural pathway to privacy preservation without sharing original data -- the key concern in cross-platform bot detection --  federated KD-based approaches baselines that can handle heterogeneity of non-IID data are the pedestal focus of the comparison.    \textbf{FedAvg} and \cite{mcmahan2017communication},
\textbf{FedProx} \cite{li2020federated} improve the local model training and update under heterogeneity through adding an optimization item.
\textbf{FedDF} \cite{lin2020ensemble} employs data-free knowledge distillation to improve the global model on server side.
\textbf{FedEnsemble} \cite{shi2021fed} uses an ensemble mechanism for combining the output of all models to predict a specific sample.
\textbf{FedDistill} \cite{shi2021fed} shares label-wise average of logit vectors among users for data-free knowledge distillation without network parameter shared.
\textbf{FedGen} \cite{zhu2021data} and \textbf{FedFTG} \cite{zhang2022fine} offer flexible parameter sharing and knowledge distillation.

\begin{figure*}[t]
\centering
\subfigure[Vendor-19 ($\alpha=1$).]{
\label{tab:heatmap-sub-1}
\includegraphics[width=5.2cm]{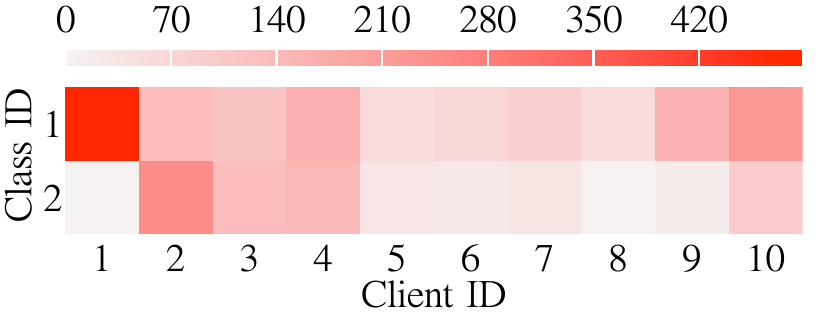}}
\hspace{3mm}
\subfigure[Vendor-19 ($\alpha=0.5$).]{
\label{tab:heatmap-sub-2}
\includegraphics[width=5.2cm]{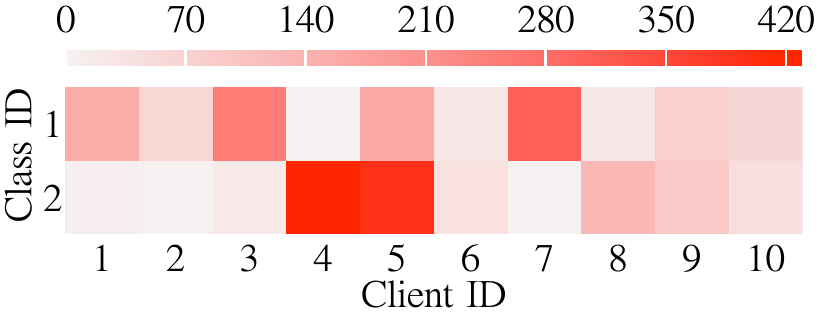}}
\hspace{3mm}
\vspace{-2mm}
\subfigure[Vendor-19 ($\alpha=0.1$).]{
\label{tab:heatmap-sub-3}
\includegraphics[width=5.2cm]{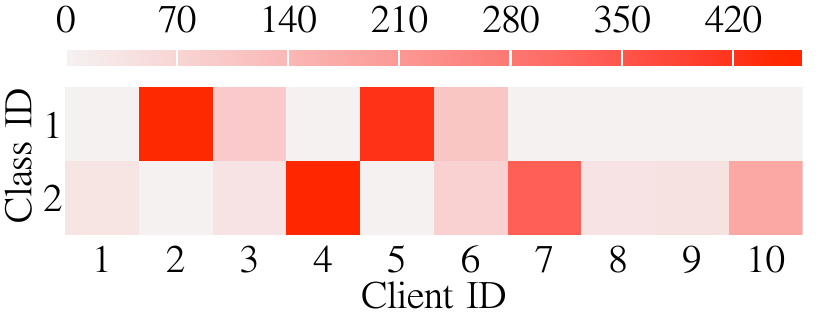}}
\subfigure[TwiBot-20 ($\alpha=1$).]{
\label{tab:heatmap-sub-4}
\includegraphics[width=5.2cm]{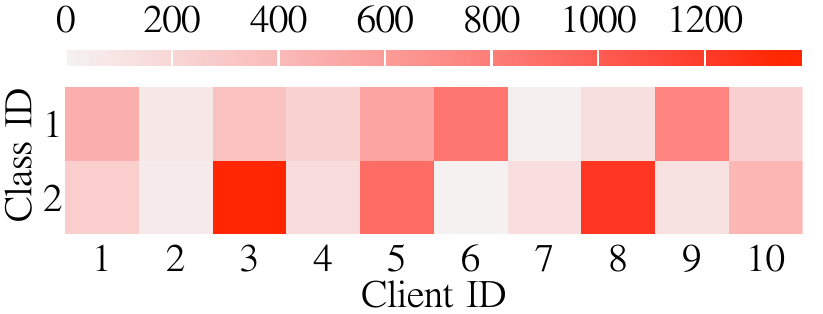}}
\hspace{3mm}
\subfigure[TwiBot-20 ($\alpha=0.5$).]{
\label{tab:heatmap-sub-5}
\includegraphics[width=5.2cm]{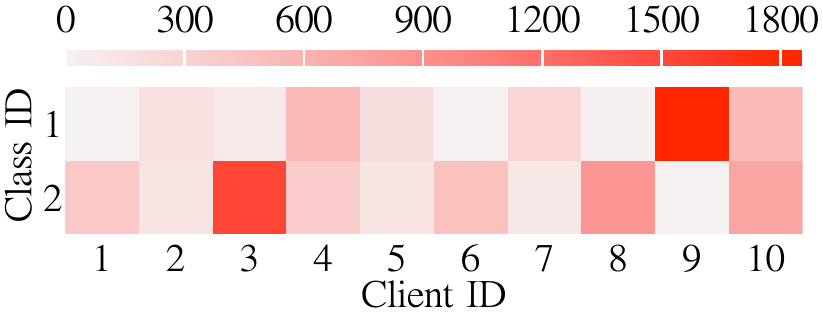}}
\hspace{3mm}
\subfigure[TwiBot-20 ($\alpha=0.1$).]{
\label{tab:heatmap-sub-6}
\includegraphics[width=5.2cm]{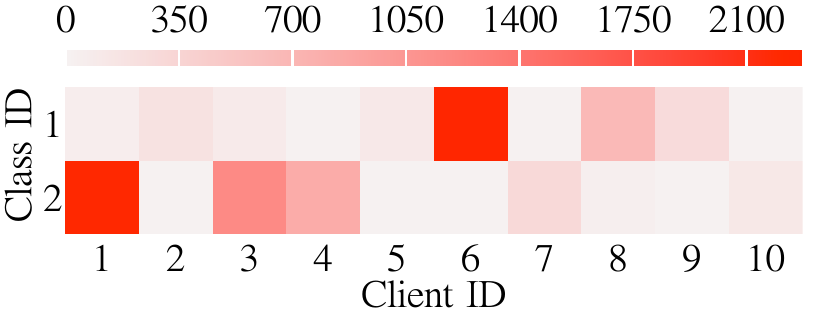}}
\vspace{-1.2em}
\caption{Visualization of data heterogeneity. The darker color means more training samples with a label available to the client.}
\label{fig:heatmap-datapartition}
\end{figure*}

\begin{table*}[tb]
\setlength{\abovecaptionskip}{0.15cm}
\setlength{\belowcaptionskip}{-0.35cm}
\caption{Comparison of the average maximum accuracy of different methods for social bot detection (\%).}
\vspace{-0.1em}
\label{tab:comparision-table}
\centering
\scalebox{0.95}{
\begin{tabular}{c|cccc|cccc}
\hline
\toprule
\specialrule{0em}{1pt}{1pt}
Dataset &\multicolumn{4}{c|}{Vendor-19} & \multicolumn{4}{c}{TwiBot-20} \\ 
\specialrule{0em}{1pt}{1pt}
\hline
\specialrule{0em}{1pt}{1pt}
Setting & $\alpha=1$  & $\alpha=0.5$  & $\alpha=0.1$ & $\alpha=0.05$ & $\alpha=1$  & $\alpha=0.5$  & $\alpha=0.1$ & $\alpha=0.05$\\
\specialrule{0em}{1pt}{1pt}
\hline
\specialrule{0em}{1pt}{1pt}
FedAvg      & 71.30$\pm$0.60    & 61.06$\pm$1.52    & 60.88$\pm$2.85    & 59.81$\pm$2.48    & 54.04$\pm$0.50    &  55.41$\pm$1.35   &  51.37$\pm$0.77   & 52.46$\pm$0.02   \\
FedProx     & 84.37$\pm$0.43    & 78.25$\pm$1.02    & 51.86$\pm$0.04    & 63.27$\pm$2.32     & 74.34$\pm$0.06    &  73.32$\pm$0.25   &  51.86$\pm$0.04   & 52.30$\pm$0.63 \\
FedDF       & 86.37$\pm$1.23    & 80.17$\pm$2.21    & 63.16$\pm$1.37    & 67.01$\pm$1.78     & 72.12$\pm$1.96    &  71.25$\pm$1.03   &  55.23$\pm$1.32   & 53.35$\pm$1.41 \\
FedEnsemble & 81.12$\pm$2.22    & 76.70$\pm$1.21    & 64.51$\pm$2.56    & 68.05$\pm$1.15     & 55.98$\pm$2.55    &  54.15$\pm$0.04   &  54.21$\pm$0.04   & 54.15$\pm$0.04   \\
FedDistill  & 79.68$\pm$0.58    & 68.77$\pm$1.13    & 52.88$\pm$0.06    & 70.25$\pm$0.39     & 64.11$\pm$0.29    &  63.34$\pm$0.56   &  50.00$\pm$0.00   & 54.30$\pm$0.05   \\
FedGen      & 90.05$\pm$0.33    & 84.83$\pm$0.96    & 65.12$\pm$0.60    & 70.79$\pm$2.39     & 74.14$\pm$0.47    &  73.12$\pm$2.09   &  59.19$\pm$2.70   & 55.78$\pm$1.79 \\
FedFTG      & 88.31$\pm$1.41    & 82.17$\pm$1.52    & 66.01$\pm$1.25    & 68.39$\pm$1.94     & 74.27$\pm$1.21    &  74.13$\pm$0.53   &  60.14 $\pm$1.74   & 56.17$\pm$1.27 \\
\specialrule{0em}{1pt}{1pt}
\hline
\specialrule{0em}{1pt}{1pt}
\FedACK-A    & \textbf{91.31$\pm$0.52}    & 84.79$\pm$1.05             & 66.10$\pm$2.90     & 68.21$\pm$1.95           & \textbf{77.16$\pm$1.09}    &  74.70$\pm$1.64   &  63.52$\pm$1.09 &\ 55.39$\pm$1.24    \\
\FedACK      & 88.58$\pm$1.91             & \textbf{87.05$\pm$2.03}      & \textbf{76.04$\pm$3.40}  & \textbf{75.27$\pm$2.50}    & 77.08$\pm$1.83             &  \textbf{78.26$\pm$2.60}                 &  \textbf{67.81$\pm$2.20}    & \textbf{60.14$\pm$1.32}          \\
\specialrule{0em}{1pt}{1pt}
\hline
\specialrule{0em}{1pt}{1pt}
Gain & \textbf{$\uparrow$\ 1.26$\sim$20.01}   & \textbf{$\uparrow$\ 2.22$\sim$25.99}     & \textbf{$\uparrow$\ 10.03$\sim$24.18}   & \textbf{$\uparrow$\ 4.48$\sim$15.46}   & \textbf{$\uparrow$\ 2.82$\sim$23.12}    &  \textbf{$\uparrow$\ 4.13$\sim$24.11}        &  \textbf{$\uparrow$\ 7.67$\sim$16.44}    &   \textbf{$\uparrow$\ 3.97$\sim$7.84}  \\

\bottomrule
\hline
\end{tabular}}
\end{table*}

\vspace{-0.2em}
\subsubsection{Model Parameters} The cross-lingual module is trained upon the corpus released in  \cite{DBLP:conf/emnlp/ZhuWWZZWZ19}. To be fair, the same cross-lingual module and backbone model $\varepsilon$ (same architecture and initial parameters) are used upon all baseline approaches.
For the transformer architectures in the cross-lingual module, we use the same configuration of \cite{vaswani2017attention}; the number of layers, feed-forward hidden size, model hidden size and the number of heads are 6, 1024, 512, and 8, respectively. The mapper is an MLP with three linear layers with a hidden dimension of 512, and the discriminator is an MLP with four linear layers with a hidden dimension of 512. The MLP used for extracting property features in the backbone model has two linear layers with a hidden dimension of 512. TextCNN \cite{DBLP:conf/emnlp/Kim14} used in $\varepsilon$  has four convolution kernels of size [2, 3, 4, 5]. The generators in \FedACK are MLP with two linear layers with a hidden dimension of 256. The discriminators in \FedACK are MLP with 3 linear layers with a hidden dimension of 256. Common parameters for training the models include: batch size (64), learning rate (0.01), optimizer (Adam), global communication rounds (100), and local updating steps (5).
 
\vspace{-0.6em}
\subsubsection{Methodology and Metrics}
The comparison is five-fold: 1) effectiveness (model accuracy and capability of handling heterogeneity), 2) efficiency (the number of communication rounds required to achieve a target accuracy) and 3) the effect of learning consistent feature space. 4) sensitivity (variation of model accuracy under different hyperparameter settings) 5) cross-lingual validation (performance gains of our proposed cross-lingual modules in cross-platform scenarios where multiple languages coexist). Since the samples of different categories in the datasets are balanced, we simply use accuracy and deviation as the main metrics.

\vspace{-1.6em}
\subsection{Effectiveness (Q1)}
\vspace{-0.1em}
We vary the hyperparameter dataset partition $\alpha$ from $\{1, 0.5, 0.1, 0.05\}$ for each dataset to validate the performance of different methods with varying degrees of data distribution heterogeneity.
The darkness of coloring represents the sample number of a specific class stored on a client. As shown in Fig.~\ref{fig:heatmap-datapartition}, increased data heterogeneity  (e.g., $\alpha=0.1$) leads to more clients store only one class of samples.

\mypara{Accuracy Comparison.} Table \ref{tab:comparision-table} compares the accuracy among baseline algorithms. All experiments are repeated over 3 random seeds.
Overall, our method outperforms all baselines in any scenario. 
\FedACK achieves 1.26\%$\sim$10.03\% accuracy improvement in absolute terms when compared with the runner-up methods (i.e., FedGen, FedFTG). While FedProx can achieve relatively competitive performance when data heterogeneity is less intense (e.g., $\alpha=\{1, 0.5\}$) due to its limit to local model updates, it cannot well handle more heterogeneous data distribution. The data-free knowledge distillation in FedGen and FedFTG can substantially improve the server's global model; however, they are insufficient to effectively tackle feature space inconsistency and model drift. The performance gain   \FedACK is more significant against other baselines when data heterogeneity increases (e.g., $\alpha=\{0.1, 0.05\}$), indicating its superiority in handling data heterogeneity. Appendix \ref{sec:append:a} further demonstrates the improved model generalization in our approach when compared with other baselines. 

\mypara{Ablation Study.} We generated a new variant model, \FedACK-A, that excludes the contrastive module. As shown in Table \ref{tab:comparision-table}, the accuracy of \FedACK-A is outstanding when data heterogeneity is low but falls off when higher heterogeneity manifests. This indicates the adversarial training and global knowledge distillation alone can function effectively in the face of low heterogeneity. The contrastive learning mechanism is of importance to constrain the model optimization direction, which demonstrate the necessity of learning consistent feature spaces when dealing with data heterogeneity.

\begin{figure*}[t]
\centering
\includegraphics[width=16cm]{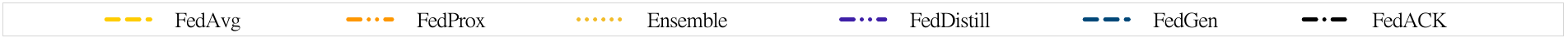}
\subfigure[Vendor-19 ($\alpha=1$).]{
\label{fig:learning-process-sub-1}
\includegraphics[width=4.31cm]{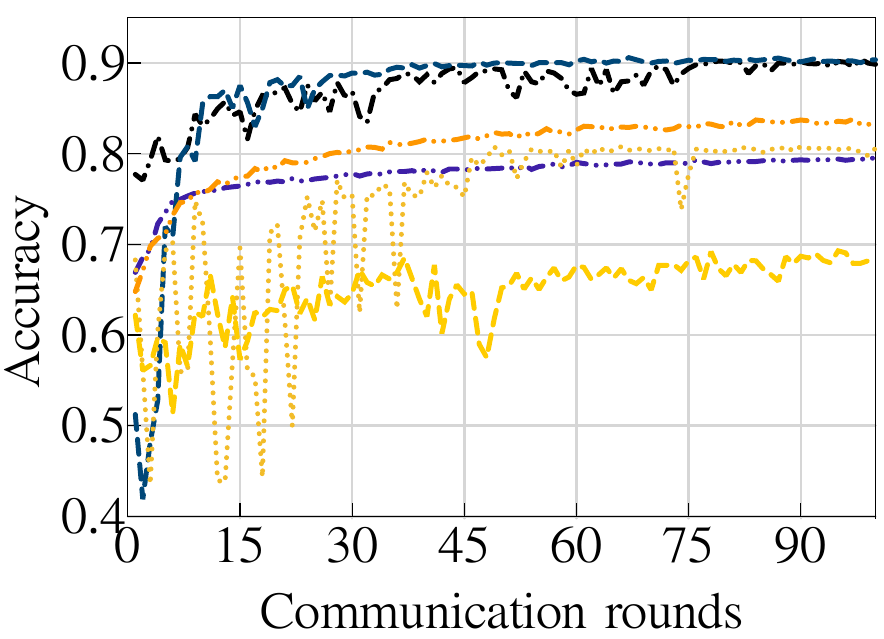}}
\subfigure[Vendor-19 ($\alpha=0.5$).]{
\label{fig:learning-process-sub-2}
\includegraphics[width=4.31cm]{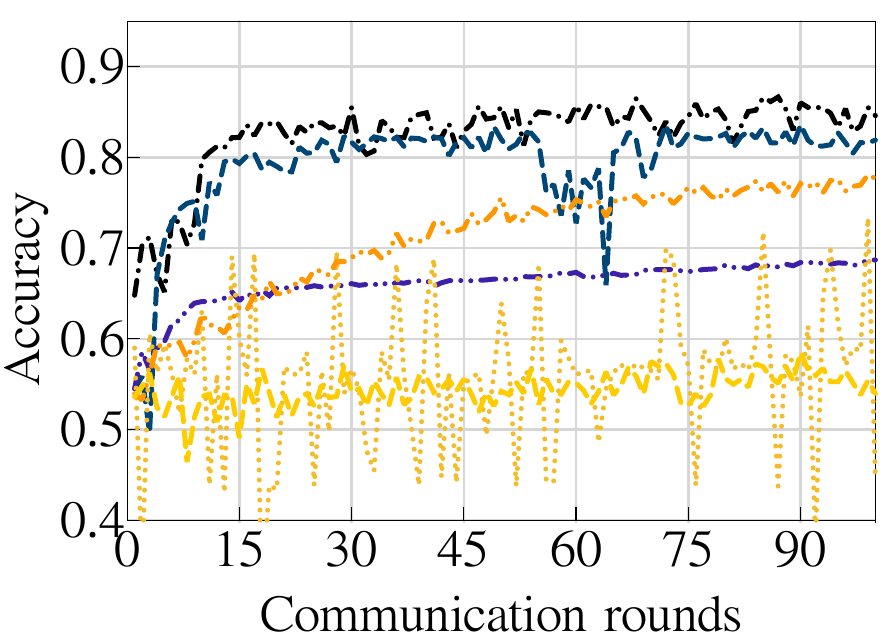}}
\subfigure[TwiBot-20 ($\alpha=1$).]{
\label{fig:learning-process-3}
\includegraphics[width=4.31cm]{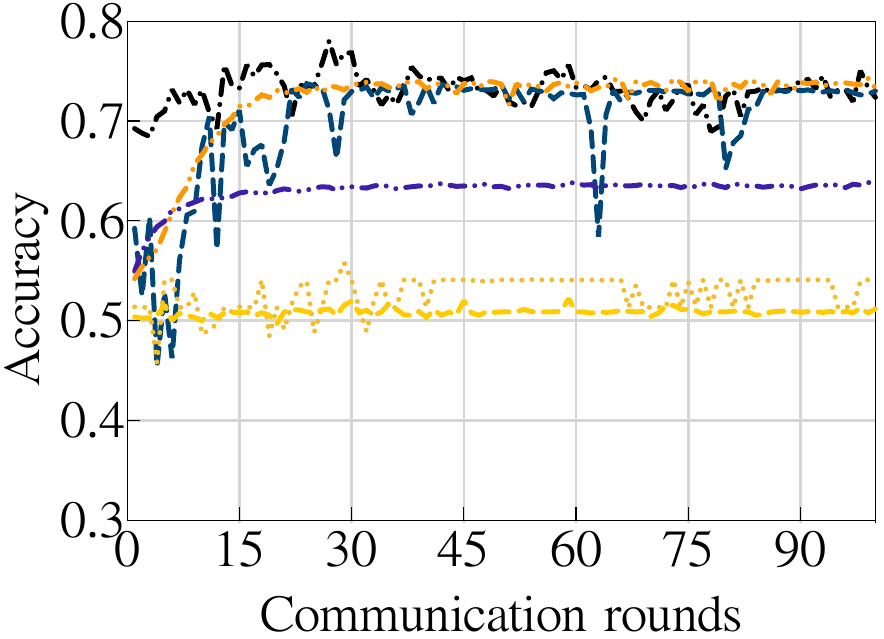}}
\subfigure[TwiBot-20 ($\alpha=0.5$).]{
\label{fig:learning-process-4}
\includegraphics[width=4.31cm]{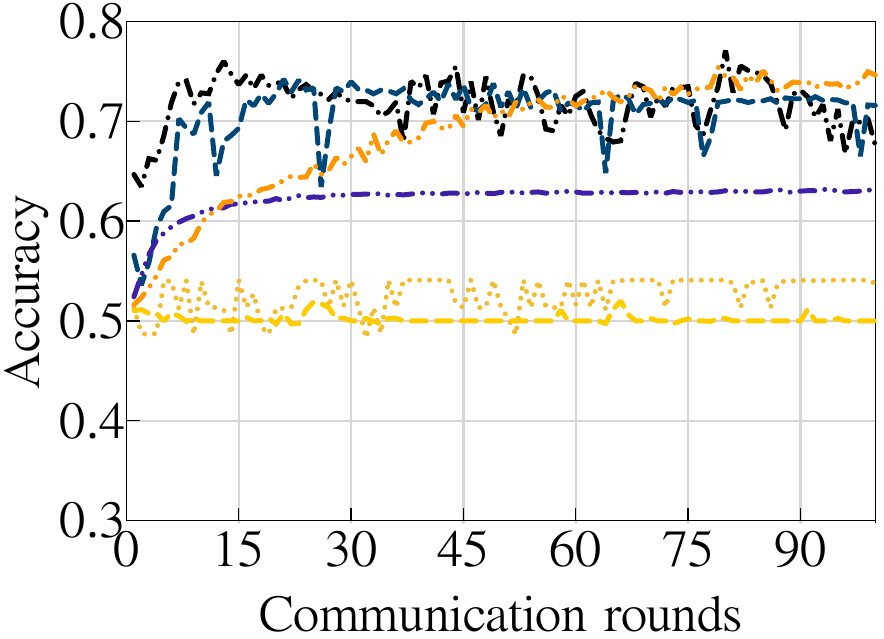}}
\vspace{-1.6em}
\caption{Learning Curve of (a-b) Vendor-19 and (c-d) TwiBot-20 in 100 communication rounds in different $\alpha$ settings.}
\label{fig:learning-process}
\end{figure*}

\begin{table}[t]
\setlength{\abovecaptionskip}{0.15cm}
\setlength{\belowcaptionskip}{-0.35cm}
\caption{The round number to reach the target accuracy on Vendor-19 $(80\%,70\%)$ and TwiBot-20 $(70\%,65\%)$.}
\vspace{-0.8mm}
\label{tab:time-consumption-table}
\centering
\scalebox{0.97}{
\begin{tabular}{@{}c|cc|cc@{}}
\hline
\toprule
Dataset             &\multicolumn{2}{c|}{Vendor-19}             & \multicolumn{2}{c}{TwiBot-20} \\ 
\specialrule{0em}{1pt}{1pt}
\hline
\specialrule{0em}{1pt}{1pt}
Setting             & $\alpha=1\ (80)$    & $\alpha=0.5\ (70)$  & $\alpha=1\ (70)$    & $\alpha=0.5\ (65)$\\
\specialrule{0em}{1pt}{1pt}
\hline
\specialrule{0em}{1pt}{1pt}
FedAvg              & unreached                     & unreached                  & unreached                  & unreached \\
FedProx             & 25.3$\pm$3.1          & 32.6$\pm$2.3      & 13.3$\pm$2.8    & 24.0$\pm$7.3  \\
FedDF             & 22.3$\pm$2.4          & 38.4$\pm$3.1      & 50.3$\pm$5.2    & 60.2$\pm$6.4  \\
Ensemble            & 9.0$\pm$1.1           & 6.0$\pm$1.4       & unreached                 & unreached      \\
FedDistill          & 60.0$\pm$1.0          & unreached                  & unreached                  & unreached \\
FedGen              & 7.3$\pm$0.4           & 5.0$\pm$0.8       & 10.6$\pm$0.9    & 4.6$\pm$1.2   \\
FedFTG              & 43.5$\pm$37.5           & 15.6$\pm$16.5       & 12.6$\pm$0.5    & 9.4$\pm$2.3   \\
\specialrule{0em}{1pt}{1pt}
\hline
\specialrule{0em}{1pt}{1pt}
\FedACK         & \textbf{4.6$\pm$3.8}      & \textbf{2.3$\pm$0.9} & \textbf{2.33$\pm$1.25} & \textbf{1.67$\pm$0.94}     \\
\bottomrule
\hline
\end{tabular}
}
\vspace{-1em}
\end{table}

\vspace{-0.2em}
\subsection{Efﬁciency (Q2)}
Fig.~\ref{fig:learning-process} shows the learning curve of different methods within 100 communication rounds and \FedACK is among the top performers. FedDistill has the best stability, rapidly approaching a stable level only after a dozen rounds of communication, but the achievable accuracy is merely lower than 0.65, making it less competitive when compared with other methods.  \FedACK can very quickly  converge to a high level of accuracy after the initial rounds and remains high in the following communication rounds. 

Table \ref{tab:time-consumption-table}  reports the average number of rounds required for each method to achieve the target accuracy in different settings. \textit{Unreached} means the failure of achieving the target accuracy ($80\%,70\%$ for Vendor-19; $70\%,65\%$ for TwiBot-20) in all three runs with different random seed. \FedACK achieves target accuracy with a minimum number of communication rounds under any circumstance. FedGen, the second best performer, still requires 1.6$\sim$4.5 times  the communication rounds of our method. This is because \FedACK incorporates the proportion of the label of each pseudo sample in each client into a part of knowledge during global knowledge extraction and classification. This indicates the importance of each client to the knowledge of a specific sample. \FedACK also limits the feature space and optimization direction of the models at the client side, resulting in quicker convergence to target a given accuracy.   

\begin{figure}[t]
\centering
\subfigure[Visualization of client 1.]{
\label{fig:boundaries-sub-1}
\includegraphics[width=0.231\textwidth]{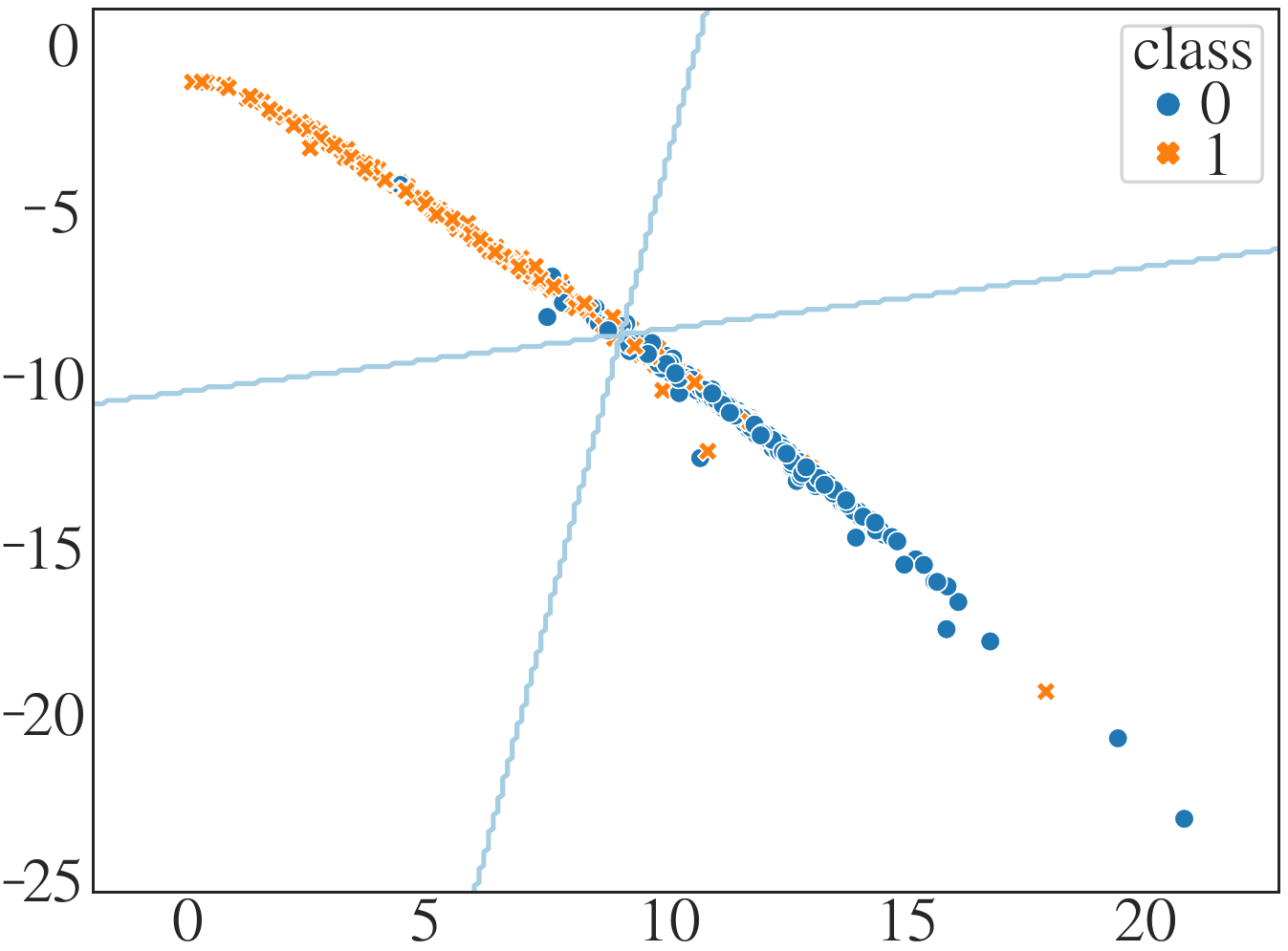}}
\subfigure[Visualization of client 2.]{
\label{fig:boundaries-sub-2}
\includegraphics[width=0.231\textwidth]{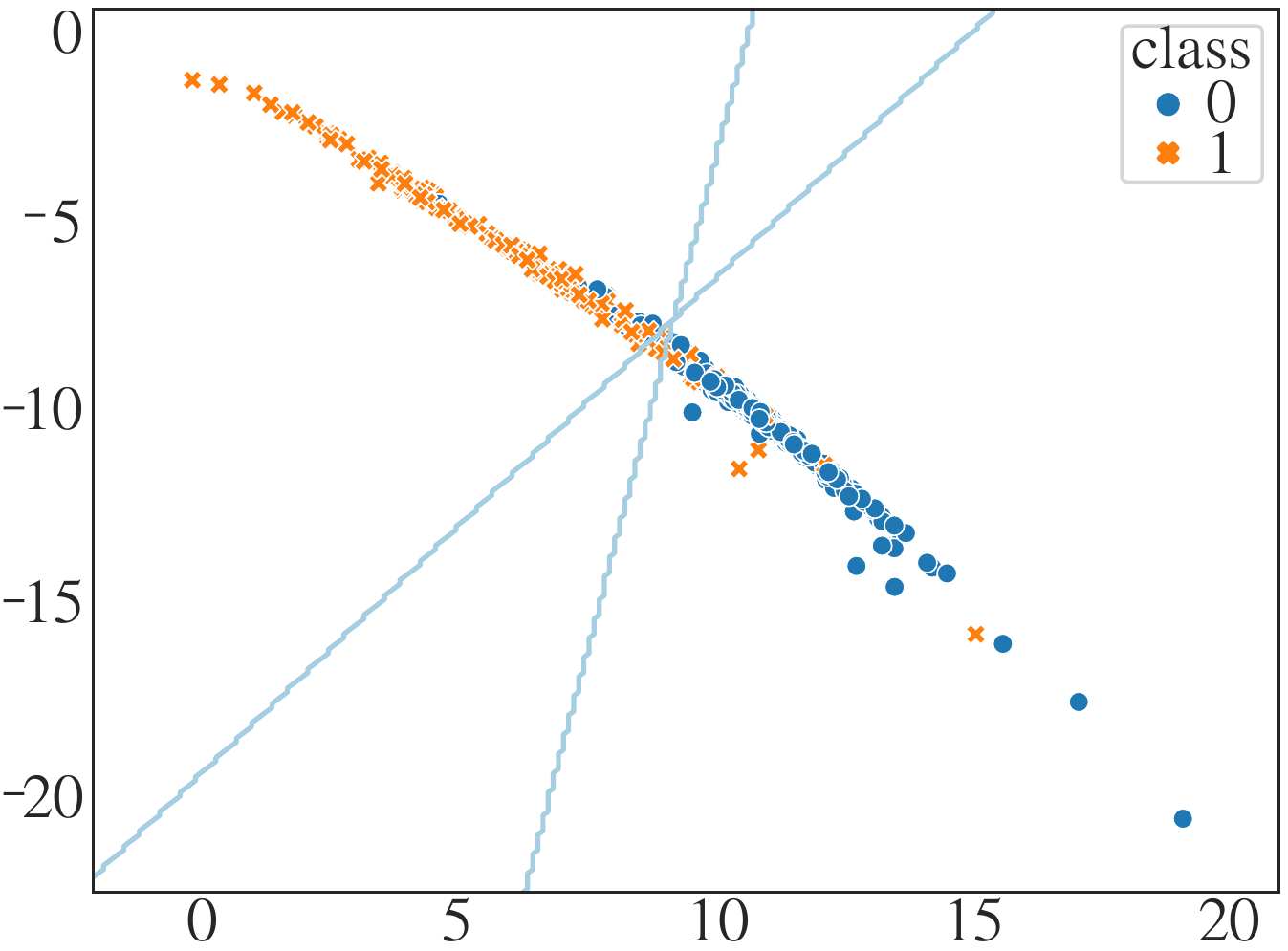}}
\vspace{-1.6em}
\caption{Decision boundaries and feature space of two randomly selected clients from \FedACK trained on Vendor-19. The x-axis and y-axis represent the values of the 2-dimensional features output by $\varepsilon$ described in Sec~\ref{exp:feature_space_consistency}.}
\vspace{-1em}
\label{fig:boundaries}
\end{figure}

\vspace{-0.1em}
\subsection{Feature Space Consistency (Q3)}
\label{exp:feature_space_consistency}

We also conduct an experiment to show how \FedACK learns the feature space. Fig.~\ref{fig:boundaries} visualizes the learnt feature space and the decision boundaries of two classifiers in \FedACK on Vendor-19 dataset. We tweak the feature extractor $\varepsilon$ to produce 2-dimension features for each input sample. We randomly select two clients after training \FedACK in 100 communication rounds and plot the features of testing data samples. 
It can be observed that adversarial learning makes the two classifiers in any of the two clients learn distinct decision boundaries. The different decision boundaries impose restrictions on the feature space learned by the feature extractor. 
To extract features from the same class of samples and locate them in overlapping areas on the same side of the decision boundary, the feature extractor compresses the generated features into a linear region for simultaneous classification. Another observation on Fig.~\ref{fig:boundaries-sub-1} and Fig.~\ref{fig:boundaries-sub-2} is that the feature extractors learn a consistent feature space across clients due to the contrastive learning that limits the update direction of the feature extractors. These findings show the advancements of \FedACK in learning feature spaces.

\begin{figure*}[t]
\centering
\subfigure[\FedACK with different $\gamma$ in adversarial learning.]{
\label{fig:sensitivity-sub-1}
\includegraphics[width=5.3cm]{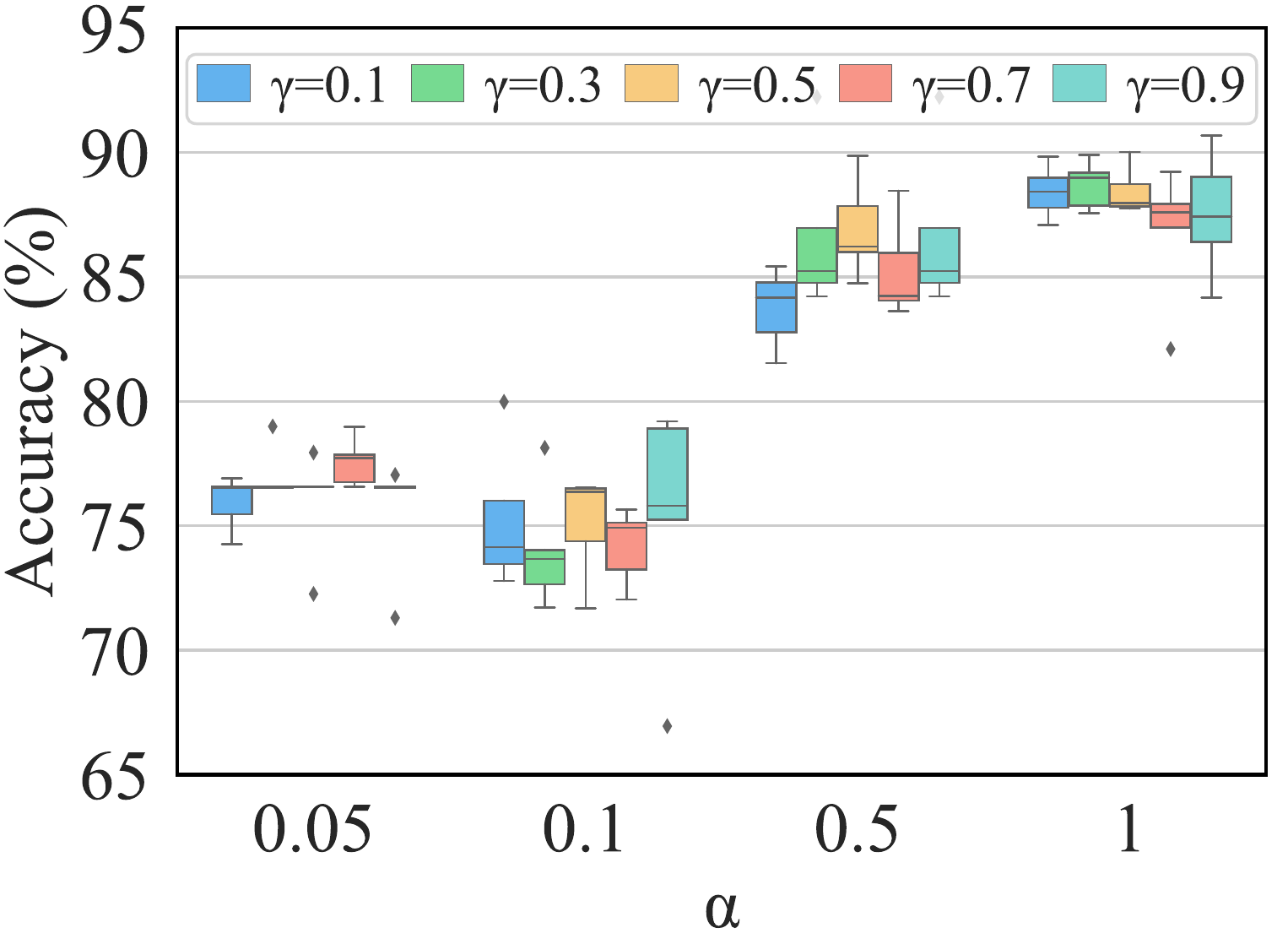}}
\hspace{2mm}
\subfigure[\FedACK with different $\mu$ in  contrastive learning.]{
\label{fig:sensitivity-sub-2}
\includegraphics[width=5.3cm]{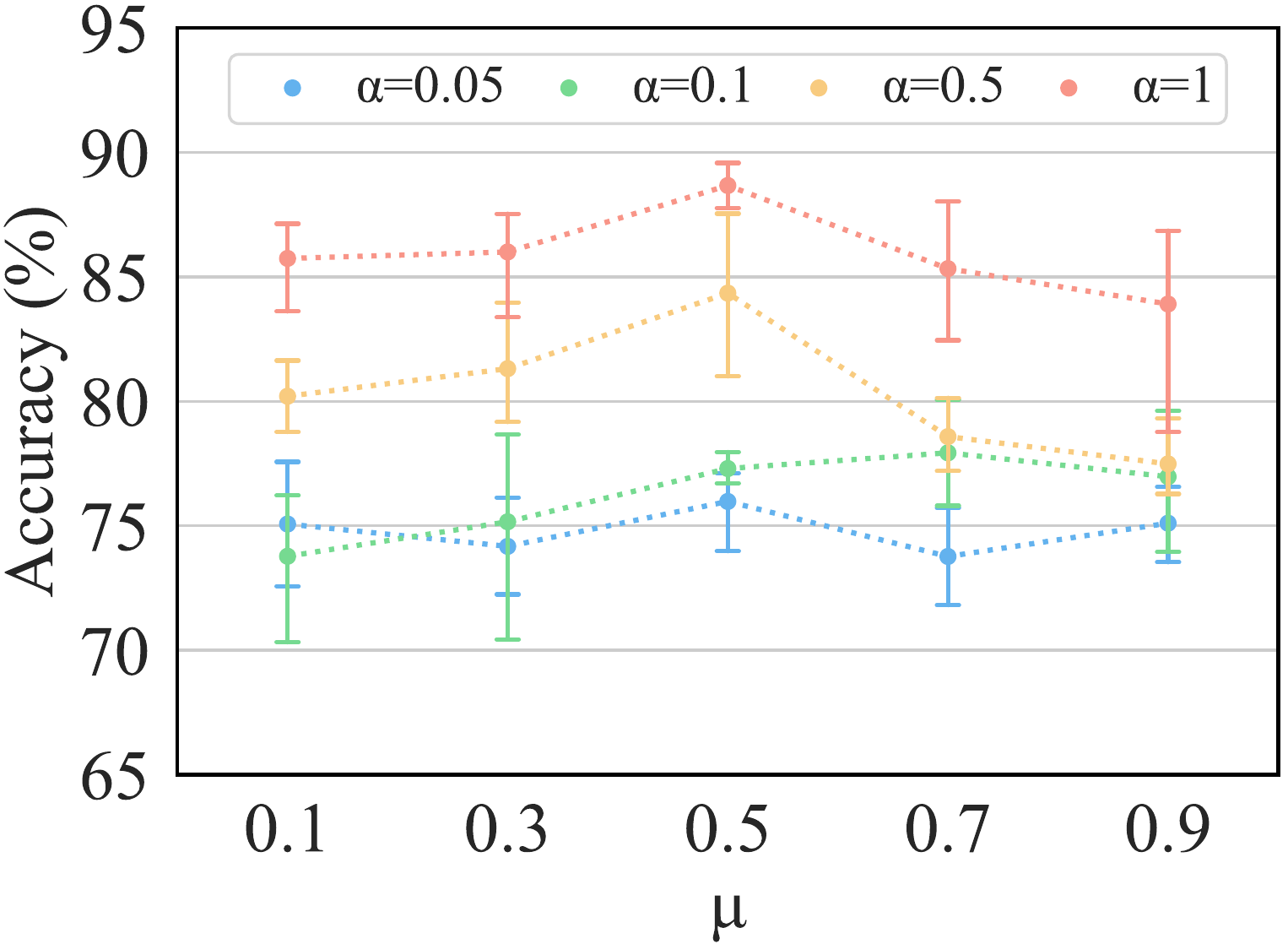}}
\hspace{2mm}
\subfigure[\FedACK with different $\tau$ in contrastive learning.]{
\label{fig:sensitivity-sub-3}
\includegraphics[width=5.3cm]{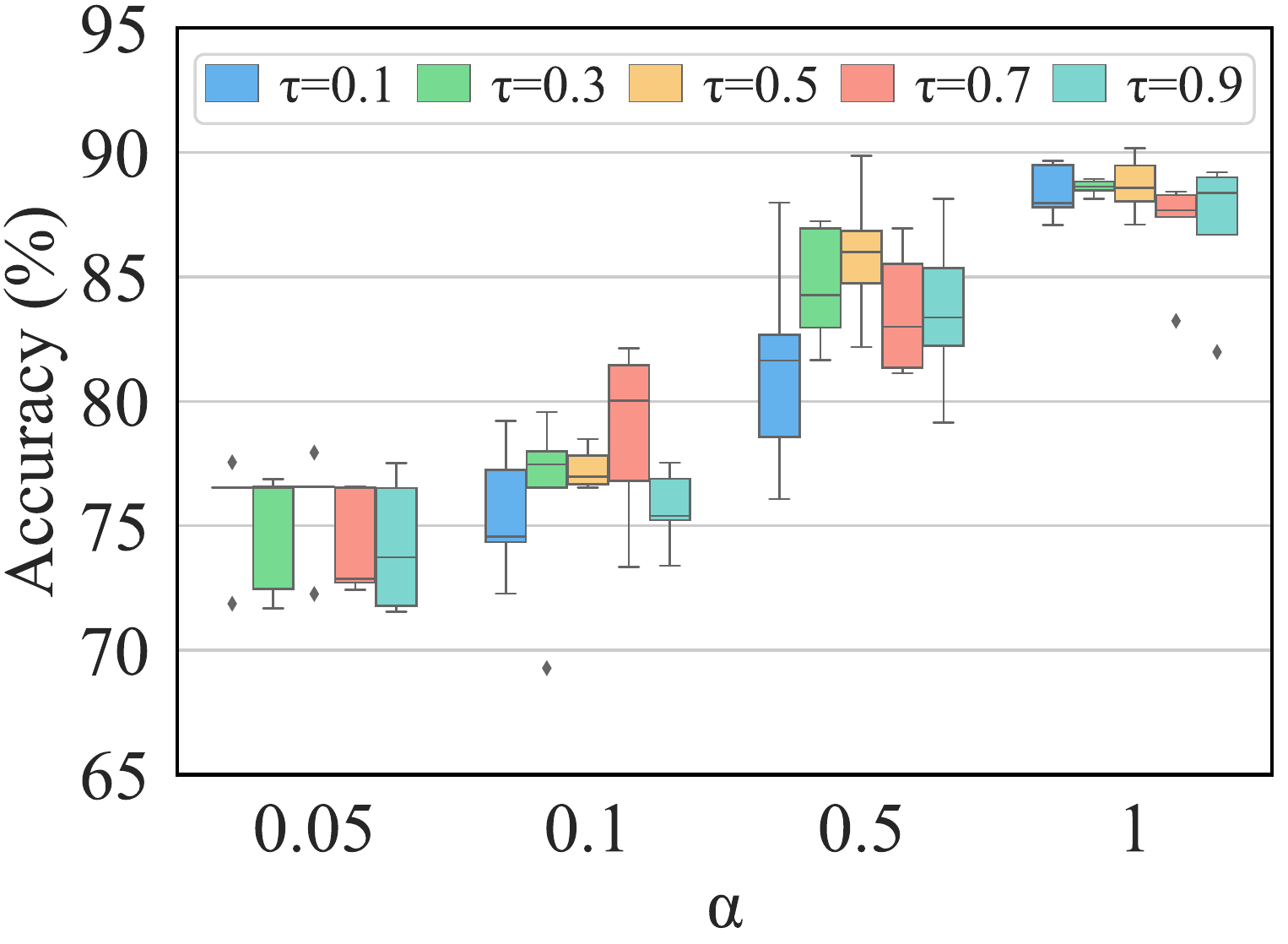}}
\vspace{-1.4em}
\caption{Hyperparameter Sensitivity ($\gamma,\mu,\tau$) of \FedACK (a-c) on Vendor-19 under different data heterogeneity settings ($\alpha$).}
\vspace{-0.5mm}
\label{fig:hyperparameter-sensitivity}
\end{figure*}

\vspace{-0.1em}
\subsection{Sensitivity (Q4)}
\vspace{-0.1em}

We investigate the hyper-parameters sensitivity based on the Vendor-19 dataset. The hyperparameters include $\gamma$ for adjusting the proportion of adversarial loss, and $\mu$ and $\tau$ for adjusting the proportion of contrastive loss. 
The number of repeated random seeds is set as 5. 

As shown in Fig.~\ref{fig:sensitivity-sub-1}, when the heterogeneity of data distribution is marginal (i.e., $\alpha = 1$), the accuracy is insensitive to $\gamma$. However, when the heterogeneity increases (i.e., $\alpha$ going down), noticeable accuracy variation manifests among different settings of $\gamma$. Similar observations can be found in Fig.~\ref{fig:sensitivity-sub-3}, indicating a discrepancy in the accuracy among different $\tau$. These findings implicate a great need for carefully examining and fine-tuning such parameters in the adversarial and contrastive loss on a case-by-case basis, considering the data characteristics (particularly the heterogeneity of data distribution), to target the optimal model performance. Fig~\ref{fig:sensitivity-sub-2} depicts the accuracy under different combinations of $\mu$ and $\alpha$. Observably, for a given data distribution, the model accuracy reaches its peak when $\mu$ increases to 0.5, before falling off if $\mu$ continues to increase. This is largely due to a balance between adversarial loss and contrastive loss. The dominance of either side will lead to a degradation of the model's effectiveness.

\vspace{-0.2em}
\subsection{Cross-Lingual Validation (Q5)}

As there is no publicly available datasets in the field of social bot detection designated for cross-lingual performance at the time of writing, we synthesize an experimental cross-lingual dataset by randomly selecting half the social accounts from Vendor-19 dataset and translating their tweets into Chinese using Google Translate. We use the same experiment settings described in Section~\ref{exp:setup}. We pre-train the cross-lingual mapper by using the cross-lingual summarization corpus NCLS \cite{DBLP:conf/emnlp/ZhuWWZZWZ19} and then combine the pre-trained encoder $\phi_E$ and mapper $\mathcal{M}$ with \FedACK and other baselines. By default, each method is evaluated with the cross-lingual mapper $\mathcal{M}$ enabled. Each method with the suffix \textit{-NC} means the model only uses $\phi_E$ to extract features from text content without $\mathcal{M}$. 

Experimental results are shown in Table~\ref{tab:cross-lingual-table}. The observations are three-fold: 
i)  \FedACK constantly outperforms others in all circumstances. The accuracy can increase 3.28$\sim$6.45 in absolute value, when data heterogeneity is low ($\alpha=1$). ii) Compared with the model variant without $\mathcal{M}$ (e.g. \FedACK-NC), the accuracy of the model (e.g. \FedACK) equipped with the cross-lingual mapper is unsurprisingly improved. This once again demonstrates the substantial capability and necessity of coping with cross-lingual issues. It is worth noting that there is also a diminishing performance gain from cross-language mapping as the $\alpha$ value gets smaller. This is because the heterogeneity of data distribution becomes the dominating challenge to accuracy, and the cross-lingual module itself is insufficient and thus brings limited benefit. iii) Equipped with the cross-lingual mapper, each method can well tackle the multi-lingual scenarios. There is marginal disparity between the model accuracy in the case of a singular language environment (shown in Table~\ref{tab:comparision-table}) and the accuracy when multiple languages co-exist in the tweet features (as shown in Table~\ref{tab:cross-lingual-table}). This demonstrates the robustness and effectiveness of the proposed mapper for cross-lingual contents. Among all the comparative baselines, our method has the least degradation caused by the existence of multiple languages, owing to the unified feature space mapping for different languages.

\begin{table}[tb]
\setlength{\abovecaptionskip}{0.15cm}
\setlength{\belowcaptionskip}{-0.35cm}
\caption{Comparison of the average maximum accuracy (\%) of different methods with/without(-NC) cross-lingual mapping. Gain is the disparity between \FedACK and other baselines.}
\label{tab:cross-lingual-table}
\centering
\scalebox{0.86}{
\begin{tabular}{c|cccc}
\hline
\toprule
\specialrule{0em}{1pt}{1pt}
Dataset &\multicolumn{4}{c}{Vendor-19} \\ 
\specialrule{0em}{1pt}{1pt}
\hline
\specialrule{0em}{1pt}{1pt}
Setting & $\alpha=1$  & $\alpha=0.5$  & $\alpha=0.1$ & $\alpha=0.05$\\
\specialrule{0em}{1pt}{1pt}
\hline
\specialrule{0em}{1pt}{1pt}
FedDistill-NC     & 78.14$\pm$1.30    & 67.01$\pm$0.16    & 66.35$\pm$0.25    & 69.24$\pm$0.30 \\
FedDistill       & 80.03$\pm$0.59    & 68.87$\pm$0.30    & 67.20$\pm$0.22    & 70.18$\pm$0.45  \\
FedGen-NC & 81.94$\pm$0.82    & 77.88$\pm$0.52    & 70.65$\pm$2.73    & 72.73$\pm$2.20  \\
FedGen  & 83.20$\pm$1.07    & 79.63$\pm$0.57    & 72.42$\pm$2.18    & 73.26$\pm$0.53  \\
FedFTG-NC      & 78.18$\pm$1.04    & 76.79$\pm$1.22    & 67.61$\pm$1.57  & 72.30$\pm$1.20 \\
FedFTG      & 82.85$\pm$1.52    & 77.29$\pm$3.42    & 69.65$\pm$0.95  & 73.15$\pm$1.27\\
\specialrule{0em}{1pt}{1pt}
\hline
\specialrule{0em}{1pt}{1pt}
\FedACK-NC    & 84.54$\pm$1.30    & 79.37$\pm$1.24             & 73.99$\pm$2.55    & \textbf{74.48$\pm$ 1.50} \\
\FedACK      & \textbf{86.48$\pm$0.99}             & \textbf{81.16$\pm$1.06}      & \textbf{75.05$\pm$2.78}  & 74.18$\pm$1.67      \\
\specialrule{0em}{1pt}{1pt}
\hline
\specialrule{0em}{1pt}{1pt}
Gain  & \textbf{$\uparrow$\ 3.28$\sim$6.45}   & \textbf{$\uparrow$\ 1.53$\sim$12.29}    & \textbf{$\uparrow$\ 2.63$\sim$7.85}  & \textbf{$\uparrow$\ 0.92$\sim$4.0} \\
\bottomrule
\hline
\end{tabular}}
\end{table}

\vspace{-0.4em}
\section{Conclusion}
Social bots have been growing in the social media platforms for years. State-of-the-art bot detection methods fail to incorporate individual platforms with different models and data characteristics. In this paper, we devise a GAN-based federated knowledge distillation mechanism for efficiently transferring knowledge of data distribution  among clients. We elaborate a contrast and adversarial learning method to ensure consistent feature space for better knowledge transfer and representation when tackling non-IID data and data scarcity among
clients. Experiments show that \FedACK outperforms baseline methods in terms of accuracy, learning efficiency, and feature space consistency. In the future, we will theoretically investigate the feature space consistency and extend \FedACK to fit graph-based datasets with diversified entities and relations.

\begin{acks}
This paper was supported by the National Key R\&D Program Program of China (2021YFC3300502). 
Hao Peng was also supported by the National Key R\&D Program of China (2021YFB1714800), NSFC through grant U20B2053, and S\&T Program of Hebei (21340301D).
\end{acks}

\bibliographystyle{ACM-Reference-Format}
\bibliography{sample-base}


\begin{thebibliography}{50}


\ifx \showCODEN    \undefined \def \showCODEN     #1{\unskip}     \fi
\ifx \showDOI      \undefined \def \showDOI       #1{#1}\fi
\ifx \showISBNx    \undefined \def \showISBNx     #1{\unskip}     \fi
\ifx \showISBNxiii \undefined \def \showISBNxiii  #1{\unskip}     \fi
\ifx \showISSN     \undefined \def \showISSN      #1{\unskip}     \fi
\ifx \showLCCN     \undefined \def \showLCCN      #1{\unskip}     \fi
\ifx \shownote     \undefined \def \shownote      #1{#1}          \fi
\ifx \showarticletitle \undefined \def \showarticletitle #1{#1}   \fi
\ifx \showURL      \undefined \def \showURL       {\relax}        \fi
\providecommand\bibfield[2]{#2}
\providecommand\bibinfo[2]{#2}
\providecommand\natexlab[1]{#1}
\providecommand\showeprint[2][]{arXiv:#2}

\bibitem[Abokhodair et~al\mbox{.}(2015)]%
        {abokhodair2015dissecting}
\bibfield{author}{\bibinfo{person}{Norah Abokhodair}, \bibinfo{person}{Daisy
  Yoo}, {and} \bibinfo{person}{David~W McDonald}.}
  \bibinfo{year}{2015}\natexlab{}.
\newblock \showarticletitle{Dissecting a social botnet: Growth, content and
  influence in Twitter}. In \bibinfo{booktitle}{\emph{CSCW}}.
  \bibinfo{pages}{839--851}.
\newblock


\bibitem[Acar et~al\mbox{.}(2021)]%
        {DBLP:conf/iclr/AcarZNMWS21}
\bibfield{author}{\bibinfo{person}{Durmus Alp~Emre Acar}, \bibinfo{person}{Yue
  Zhao}, \bibinfo{person}{Ramon~Matas Navarro}, \bibinfo{person}{Matthew
  Mattina}, \bibinfo{person}{Paul~N. Whatmough}, {and}
  \bibinfo{person}{Venkatesh Saligrama}.} \bibinfo{year}{2021}\natexlab{}.
\newblock \showarticletitle{Federated Learning Based on Dynamic
  Regularization}. In \bibinfo{booktitle}{\emph{ICLR}}.
  \bibinfo{publisher}{OpenReview.net}.
\newblock


\bibitem[Berger and Morgan(2015)]%
        {berger2015isis}
\bibfield{author}{\bibinfo{person}{Jonathon~M Berger} {and}
  \bibinfo{person}{Jonathon Morgan}.} \bibinfo{year}{2015}\natexlab{}.
\newblock \showarticletitle{The ISIS Twitter Census: Defining and describing
  the population of ISIS supporters on Twitter}.
\newblock  (\bibinfo{year}{2015}).
\newblock


\bibitem[Bonawitz et~al\mbox{.}(2017)]%
        {bonawitz2017practical}
\bibfield{author}{\bibinfo{person}{Keith Bonawitz}, \bibinfo{person}{Vladimir
  Ivanov}, \bibinfo{person}{Ben Kreuter}, \bibinfo{person}{Antonio Marcedone},
  \bibinfo{person}{H~Brendan McMahan}, \bibinfo{person}{Sarvar Patel},
  \bibinfo{person}{Daniel Ramage}, \bibinfo{person}{Aaron Segal}, {and}
  \bibinfo{person}{Karn Seth}.} \bibinfo{year}{2017}\natexlab{}.
\newblock \showarticletitle{Practical secure aggregation for privacy-preserving
  machine learning}. In \bibinfo{booktitle}{\emph{CCS}}.
  \bibinfo{pages}{1175--1191}.
\newblock


\bibitem[Buciluǎ et~al\mbox{.}(2006)]%
        {bucilu2006model}
\bibfield{author}{\bibinfo{person}{Cristian Buciluǎ}, \bibinfo{person}{Rich
  Caruana}, {and} \bibinfo{person}{Alexandru Niculescu-Mizil}.}
  \bibinfo{year}{2006}\natexlab{}.
\newblock \showarticletitle{Model compression}. In
  \bibinfo{booktitle}{\emph{SIGKDD}}. \bibinfo{pages}{535--541}.
\newblock


\bibitem[Chi et~al\mbox{.}(2020)]%
        {chi2020infoxlm}
\bibfield{author}{\bibinfo{person}{Zewen Chi}, \bibinfo{person}{Li Dong},
  \bibinfo{person}{Furu Wei}, \bibinfo{person}{Nan Yang},
  \bibinfo{person}{Saksham Singhal}, \bibinfo{person}{Wenhui Wang},
  \bibinfo{person}{Xia Song}, \bibinfo{person}{Xian-Ling Mao},
  \bibinfo{person}{Heyan Huang}, {and} \bibinfo{person}{Ming Zhou}.}
  \bibinfo{year}{2020}\natexlab{}.
\newblock \showarticletitle{InfoXLM: An information-theoretic framework for
  cross-lingual language model pre-training}.
\newblock \bibinfo{journal}{\emph{arXiv preprint arXiv:2007.07834}}
  (\bibinfo{year}{2020}).
\newblock


\bibitem[Chu et~al\mbox{.}(2021)]%
        {chu2021cross}
\bibfield{author}{\bibinfo{person}{Samuel Kai~Wah Chu}, \bibinfo{person}{Runbin
  Xie}, {and} \bibinfo{person}{Yanshu Wang}.} \bibinfo{year}{2021}\natexlab{}.
\newblock \showarticletitle{Cross-Language fake news detection}.
\newblock \bibinfo{journal}{\emph{DIM}} \bibinfo{volume}{5},
  \bibinfo{number}{1} (\bibinfo{year}{2021}), \bibinfo{pages}{100--109}.
\newblock


\bibitem[Cresci(2020)]%
        {DBLP:journals/cacm/Cresci20}
\bibfield{author}{\bibinfo{person}{Stefano Cresci}.}
  \bibinfo{year}{2020}\natexlab{}.
\newblock \showarticletitle{A decade of social bot detection}.
\newblock \bibinfo{journal}{\emph{Commun. {ACM}}} \bibinfo{volume}{63},
  \bibinfo{number}{10} (\bibinfo{year}{2020}), \bibinfo{pages}{72--83}.
\newblock


\bibitem[D'Andrea et~al\mbox{.}(2015)]%
        {d2015real}
\bibfield{author}{\bibinfo{person}{Eleonora D'Andrea}, \bibinfo{person}{Pietro
  Ducange}, \bibinfo{person}{Beatrice Lazzerini}, {and}
  \bibinfo{person}{Francesco Marcelloni}.} \bibinfo{year}{2015}\natexlab{}.
\newblock \showarticletitle{Real-time detection of traffic from twitter stream
  analysis}.
\newblock \bibinfo{journal}{\emph{T-ITS}} \bibinfo{volume}{16},
  \bibinfo{number}{4} (\bibinfo{year}{2015}), \bibinfo{pages}{2269--2283}.
\newblock


\bibitem[De et~al\mbox{.}(2022)]%
        {DBLP:journals/talip/DeBGE22}
\bibfield{author}{\bibinfo{person}{Arkadipta De}, \bibinfo{person}{Dibyanayan
  Bandyopadhyay}, \bibinfo{person}{Baban Gain}, {and} \bibinfo{person}{Asif
  Ekbal}.} \bibinfo{year}{2022}\natexlab{}.
\newblock \showarticletitle{A Transformer-Based Approach to Multilingual Fake
  News Detection in Low-Resource Languages}.
\newblock \bibinfo{journal}{\emph{{ACM} Trans. Asian Low Resour. Lang. Inf.
  Process.}} \bibinfo{volume}{21}, \bibinfo{number}{1} (\bibinfo{year}{2022}),
  \bibinfo{pages}{9:1--9:20}.
\newblock


\bibitem[Deb et~al\mbox{.}(2019)]%
        {deb2019perils}
\bibfield{author}{\bibinfo{person}{Ashok Deb}, \bibinfo{person}{Luca Luceri},
  \bibinfo{person}{Adam Badaway}, {and} \bibinfo{person}{Emilio Ferrara}.}
  \bibinfo{year}{2019}\natexlab{}.
\newblock \showarticletitle{Perils and challenges of social media and election
  manipulation analysis: The 2018 us midterms}. In
  \bibinfo{booktitle}{\emph{WWW}}. \bibinfo{pages}{237--247}.
\newblock


\bibitem[Dementieva and Panchenko(2021)]%
        {dementieva2021cross}
\bibfield{author}{\bibinfo{person}{Daryna Dementieva} {and}
  \bibinfo{person}{Alexander Panchenko}.} \bibinfo{year}{2021}\natexlab{}.
\newblock \showarticletitle{Cross-lingual Evidence Improves Monolingual Fake
  News Detection}. In \bibinfo{booktitle}{\emph{ACL(Workshop)}}.
  \bibinfo{pages}{310--320}.
\newblock


\bibitem[Du et~al\mbox{.}(2021)]%
        {du2021cross}
\bibfield{author}{\bibinfo{person}{Jiangshu Du}, \bibinfo{person}{Yingtong
  Dou}, \bibinfo{person}{Congying Xia}, \bibinfo{person}{Limeng Cui},
  \bibinfo{person}{Jing Ma}, {and} \bibinfo{person}{S~Yu Philip}.}
  \bibinfo{year}{2021}\natexlab{}.
\newblock \showarticletitle{Cross-lingual covid-19 fake news detection}. In
  \bibinfo{booktitle}{\emph{ICDMW (Workshops)}}. IEEE,
  \bibinfo{pages}{859--862}.
\newblock


\bibitem[Feng et~al\mbox{.}(2022)]%
        {DBLP:journals/corr/abs-2109-02927}
\bibfield{author}{\bibinfo{person}{Shangbin Feng}, \bibinfo{person}{Zhaoxuan
  Tan}, \bibinfo{person}{Rui Li}, {and} \bibinfo{person}{Minnan Luo}.}
  \bibinfo{year}{2022}\natexlab{}.
\newblock \showarticletitle{Heterogeneity-aware Twitter Bot Detection with
  Relational Graph Transformers}. In \bibinfo{booktitle}{\emph{AAAI}}.
\newblock


\bibitem[Feng et~al\mbox{.}(2021a)]%
        {feng2021satar}
\bibfield{author}{\bibinfo{person}{Shangbin Feng}, \bibinfo{person}{Herun Wan},
  \bibinfo{person}{Ningnan Wang}, \bibinfo{person}{Jundong Li}, {and}
  \bibinfo{person}{Minnan Luo}.} \bibinfo{year}{2021}\natexlab{a}.
\newblock \showarticletitle{Satar: A self-supervised approach to twitter
  account representation learning and its application in bot detection}. In
  \bibinfo{booktitle}{\emph{CIKM}}. \bibinfo{pages}{3808--3817}.
\newblock


\bibitem[Feng et~al\mbox{.}(2021b)]%
        {feng2021twibot}
\bibfield{author}{\bibinfo{person}{Shangbin Feng}, \bibinfo{person}{Herun Wan},
  \bibinfo{person}{Ningnan Wang}, \bibinfo{person}{Jundong Li}, {and}
  \bibinfo{person}{Minnan Luo}.} \bibinfo{year}{2021}\natexlab{b}.
\newblock \showarticletitle{Twibot-20: A comprehensive twitter bot detection
  benchmark}. In \bibinfo{booktitle}{\emph{CIKM}}. \bibinfo{pages}{4485--4494}.
\newblock


\bibitem[Ferrara et~al\mbox{.}(2020)]%
        {ferrara2020characterizing}
\bibfield{author}{\bibinfo{person}{Emilio Ferrara}, \bibinfo{person}{Herbert
  Chang}, \bibinfo{person}{Emily Chen}, \bibinfo{person}{Goran Muric}, {and}
  \bibinfo{person}{Jaimin Patel}.} \bibinfo{year}{2020}\natexlab{}.
\newblock \showarticletitle{Characterizing social media manipulation in the
  2020 US presidential election}.
\newblock \bibinfo{journal}{\emph{First Monday}} (\bibinfo{year}{2020}).
\newblock


\bibitem[Ferrara et~al\mbox{.}(2016)]%
        {ferrara2016predicting}
\bibfield{author}{\bibinfo{person}{Emilio Ferrara}, \bibinfo{person}{Wen-Qiang
  Wang}, \bibinfo{person}{Onur Varol}, \bibinfo{person}{Alessandro Flammini},
  {and} \bibinfo{person}{Aram Galstyan}.} \bibinfo{year}{2016}\natexlab{}.
\newblock \showarticletitle{Predicting online extremism, content adopters, and
  interaction reciprocity}. In \bibinfo{booktitle}{\emph{ICSI}}. Springer,
  \bibinfo{pages}{22--39}.
\newblock


\bibitem[Hinton et~al\mbox{.}(2015)]%
        {hinton2015distilling}
\bibfield{author}{\bibinfo{person}{Geoffrey Hinton}, \bibinfo{person}{Oriol
  Vinyals}, \bibinfo{person}{Jeff Dean}, {et~al\mbox{.}}}
  \bibinfo{year}{2015}\natexlab{}.
\newblock \showarticletitle{Distilling the knowledge in a neural network}.
\newblock \bibinfo{journal}{\emph{arXiv}} \bibinfo{volume}{2},
  \bibinfo{number}{7} (\bibinfo{year}{2015}).
\newblock


\bibitem[Karimireddy et~al\mbox{.}(2020)]%
        {karimireddy2020scaffold}
\bibfield{author}{\bibinfo{person}{Sai~Praneeth Karimireddy},
  \bibinfo{person}{Satyen Kale}, \bibinfo{person}{Mehryar Mohri},
  \bibinfo{person}{Sashank Reddi}, \bibinfo{person}{Sebastian Stich}, {and}
  \bibinfo{person}{Ananda~Theertha Suresh}.} \bibinfo{year}{2020}\natexlab{}.
\newblock \showarticletitle{Scaffold: Stochastic controlled averaging for
  federated learning}. In \bibinfo{booktitle}{\emph{ICML}}. PMLR,
  \bibinfo{pages}{5132--5143}.
\newblock


\bibitem[Kim(2014)]%
        {DBLP:conf/emnlp/Kim14}
\bibfield{author}{\bibinfo{person}{Yoon Kim}.} \bibinfo{year}{2014}\natexlab{}.
\newblock \showarticletitle{Convolutional Neural Networks for Sentence
  Classification}. In \bibinfo{booktitle}{\emph{EMNLP}}.
  \bibinfo{publisher}{ACL}, \bibinfo{pages}{1746--1751}.
\newblock


\bibitem[Kone{\v{c}}n{\'y} et~al\mbox{.}(2016)]%
        {DBLP:journals/corr/KonecnyMRR16}
\bibfield{author}{\bibinfo{person}{Jakub Kone{\v{c}}n{\'y}},
  \bibinfo{person}{H.~Brendan McMahan}, \bibinfo{person}{Daniel Ramage}, {and}
  \bibinfo{person}{Peter Richt{\'{a}}rik}.} \bibinfo{year}{2016}\natexlab{}.
\newblock \showarticletitle{Federated Optimization: Distributed Machine
  Learning for On-Device Intelligence}.
\newblock \bibinfo{journal}{\emph{CoRR}}  \bibinfo{volume}{abs/1610.02527}
  (\bibinfo{year}{2016}).
\newblock


\bibitem[Li et~al\mbox{.}(2021a)]%
        {li2021federated}
\bibfield{author}{\bibinfo{person}{Qinbin Li}, \bibinfo{person}{Yiqun Diao},
  \bibinfo{person}{Quan Chen}, {and} \bibinfo{person}{Bingsheng He}.}
  \bibinfo{year}{2021}\natexlab{a}.
\newblock \showarticletitle{Federated learning on non-iid data silos: An
  experimental study}.
\newblock \bibinfo{journal}{\emph{arXiv}} (\bibinfo{year}{2021}).
\newblock


\bibitem[Li et~al\mbox{.}(2021b)]%
        {li2021model}
\bibfield{author}{\bibinfo{person}{Qinbin Li}, \bibinfo{person}{Bingsheng He},
  {and} \bibinfo{person}{Dawn Song}.} \bibinfo{year}{2021}\natexlab{b}.
\newblock \showarticletitle{Model-contrastive federated learning}. In
  \bibinfo{booktitle}{\emph{CVPR}}. \bibinfo{pages}{10713--10722}.
\newblock


\bibitem[Li et~al\mbox{.}(2020)]%
        {li2020federated}
\bibfield{author}{\bibinfo{person}{Tian Li}, \bibinfo{person}{Anit~Kumar Sahu},
  \bibinfo{person}{Manzil Zaheer}, \bibinfo{person}{Maziar Sanjabi},
  \bibinfo{person}{Ameet Talwalkar}, {and} \bibinfo{person}{Virginia Smith}.}
  \bibinfo{year}{2020}\natexlab{}.
\newblock \showarticletitle{Federated optimization in heterogeneous networks}.
\newblock \bibinfo{journal}{\emph{PMLS}}  \bibinfo{volume}{2}
  (\bibinfo{year}{2020}), \bibinfo{pages}{429--450}.
\newblock


\bibitem[Lin et~al\mbox{.}(2020)]%
        {lin2020ensemble}
\bibfield{author}{\bibinfo{person}{Tao Lin}, \bibinfo{person}{Lingjing Kong},
  \bibinfo{person}{Sebastian~U Stich}, {and} \bibinfo{person}{Martin Jaggi}.}
  \bibinfo{year}{2020}\natexlab{}.
\newblock \showarticletitle{Ensemble distillation for robust model fusion in
  federated learning}.
\newblock \bibinfo{journal}{\emph{NIPS}}  \bibinfo{volume}{33}
  (\bibinfo{year}{2020}), \bibinfo{pages}{2351--2363}.
\newblock


\bibitem[Liu et~al\mbox{.}(2022)]%
        {liu2022federated}
\bibfield{author}{\bibinfo{person}{Zhiwei Liu}, \bibinfo{person}{Liangwei
  Yang}, \bibinfo{person}{Ziwei Fan}, \bibinfo{person}{Hao Peng}, {and}
  \bibinfo{person}{Philip~S Yu}.} \bibinfo{year}{2022}\natexlab{}.
\newblock \showarticletitle{Federated social recommendation with graph neural
  network}.
\newblock \bibinfo{journal}{\emph{TIST}} \bibinfo{volume}{13},
  \bibinfo{number}{4} (\bibinfo{year}{2022}), \bibinfo{pages}{1--24}.
\newblock


\bibitem[McMahan et~al\mbox{.}(2017)]%
        {mcmahan2017communication}
\bibfield{author}{\bibinfo{person}{Brendan McMahan}, \bibinfo{person}{Eider
  Moore}, \bibinfo{person}{Daniel Ramage}, \bibinfo{person}{Seth Hampson},
  {and} \bibinfo{person}{Blaise~Aguera y Arcas}.}
  \bibinfo{year}{2017}\natexlab{}.
\newblock \showarticletitle{Communication-efficient learning of deep networks
  from decentralized data}. In \bibinfo{booktitle}{\emph{AISTATS}}. PMLR,
  \bibinfo{pages}{1273--1282}.
\newblock


\bibitem[Nozza(2021)]%
        {DBLP:conf/acl/Nozza20}
\bibfield{author}{\bibinfo{person}{Debora Nozza}.}
  \bibinfo{year}{2021}\natexlab{}.
\newblock \showarticletitle{Exposing the limits of Zero-shot Cross-lingual Hate
  Speech Detection}. In \bibinfo{booktitle}{\emph{ACL/IJCNLP}}.
  \bibinfo{publisher}{ACL}, \bibinfo{pages}{907--914}.
\newblock


\bibitem[Peng et~al\mbox{.}(2021)]%
        {peng2021differentially}
\bibfield{author}{\bibinfo{person}{Hao Peng}, \bibinfo{person}{Haoran Li},
  \bibinfo{person}{Yangqiu Song}, \bibinfo{person}{Vincent Zheng}, {and}
  \bibinfo{person}{Jianxin Li}.} \bibinfo{year}{2021}\natexlab{}.
\newblock \showarticletitle{Differentially private federated knowledge graphs
  embedding}. In \bibinfo{booktitle}{\emph{CIKM}}. \bibinfo{pages}{1416--1425}.
\newblock


\bibitem[Peng et~al\mbox{.}(2022)]%
        {peng2022reinforced}
\bibfield{author}{\bibinfo{person}{Hao Peng}, \bibinfo{person}{Ruitong Zhang},
  \bibinfo{person}{Shaoning Li}, \bibinfo{person}{Yuwei Cao},
  \bibinfo{person}{Shirui Pan}, {and} \bibinfo{person}{Philip Yu}.}
  \bibinfo{year}{2022}\natexlab{}.
\newblock \showarticletitle{Reinforced, incremental and cross-lingual event
  detection from social messages}.
\newblock \bibinfo{journal}{\emph{TPAMI}} (\bibinfo{year}{2022}).
\newblock


\bibitem[Rasouli et~al\mbox{.}(2020)]%
        {rasouli2020fedgan}
\bibfield{author}{\bibinfo{person}{Mohammad Rasouli}, \bibinfo{person}{Tao
  Sun}, {and} \bibinfo{person}{Ram Rajagopal}.}
  \bibinfo{year}{2020}\natexlab{}.
\newblock \showarticletitle{Fedgan: Federated generative adversarial networks
  for distributed data}.
\newblock \bibinfo{journal}{\emph{arXiv}} (\bibinfo{year}{2020}).
\newblock


\bibitem[Seo et~al\mbox{.}(2020)]%
        {seo2020federated}
\bibfield{author}{\bibinfo{person}{Hyowoon Seo}, \bibinfo{person}{Jihong Park},
  \bibinfo{person}{Seungeun Oh}, \bibinfo{person}{Mehdi Bennis}, {and}
  \bibinfo{person}{Seong-Lyun Kim}.} \bibinfo{year}{2020}\natexlab{}.
\newblock \showarticletitle{Federated knowledge distillation}.
\newblock \bibinfo{journal}{\emph{arXiv}} (\bibinfo{year}{2020}).
\newblock


\bibitem[Shi et~al\mbox{.}(2021)]%
        {shi2021fed}
\bibfield{author}{\bibinfo{person}{Naichen Shi}, \bibinfo{person}{Fan Lai},
  \bibinfo{person}{Raed~Al Kontar}, {and} \bibinfo{person}{Mosharaf
  Chowdhury}.} \bibinfo{year}{2021}\natexlab{}.
\newblock \showarticletitle{Fed-ensemble: Improving generalization through
  model ensembling in federated learning}.
\newblock \bibinfo{journal}{\emph{arXiv}} (\bibinfo{year}{2021}).
\newblock


\bibitem[Singh and Jaggi(2020)]%
        {singh2020model}
\bibfield{author}{\bibinfo{person}{Sidak~Pal Singh} {and}
  \bibinfo{person}{Martin Jaggi}.} \bibinfo{year}{2020}\natexlab{}.
\newblock \showarticletitle{Model fusion via optimal transport}.
\newblock \bibinfo{journal}{\emph{NIPS}}  \bibinfo{volume}{33}
  (\bibinfo{year}{2020}), \bibinfo{pages}{22045--22055}.
\newblock


\bibitem[Steimel et~al\mbox{.}(2019)]%
        {steimel2019investigating}
\bibfield{author}{\bibinfo{person}{Kenneth Steimel}, \bibinfo{person}{Daniel
  Dakota}, \bibinfo{person}{Yue Chen}, {and} \bibinfo{person}{Sandra
  K{\"u}bler}.} \bibinfo{year}{2019}\natexlab{}.
\newblock \showarticletitle{Investigating multilingual abusive language
  detection: A cautionary tale}. In \bibinfo{booktitle}{\emph{RANLP}}.
  \bibinfo{pages}{1151--1160}.
\newblock


\bibitem[Varol et~al\mbox{.}(2017)]%
        {varol2017online}
\bibfield{author}{\bibinfo{person}{Onur Varol}, \bibinfo{person}{Emilio
  Ferrara}, \bibinfo{person}{Clayton Davis}, \bibinfo{person}{Filippo Menczer},
  {and} \bibinfo{person}{Alessandro Flammini}.}
  \bibinfo{year}{2017}\natexlab{}.
\newblock \showarticletitle{Online human-bot interactions: Detection,
  estimation, and characterization}. In \bibinfo{booktitle}{\emph{ICWSM}},
  Vol.~\bibinfo{volume}{11}. \bibinfo{pages}{280--289}.
\newblock


\bibitem[Vaswani et~al\mbox{.}(2017)]%
        {vaswani2017attention}
\bibfield{author}{\bibinfo{person}{Ashish Vaswani}, \bibinfo{person}{Noam
  Shazeer}, \bibinfo{person}{Niki Parmar}, \bibinfo{person}{Jakob Uszkoreit},
  \bibinfo{person}{Llion Jones}, \bibinfo{person}{Aidan~N Gomez},
  \bibinfo{person}{{\L}ukasz Kaiser}, {and} \bibinfo{person}{Illia
  Polosukhin}.} \bibinfo{year}{2017}\natexlab{}.
\newblock \showarticletitle{Attention is all you need}.
\newblock \bibinfo{journal}{\emph{NIPS}}  \bibinfo{volume}{30}
  (\bibinfo{year}{2017}).
\newblock


\bibitem[Wei and Nguyen(2019)]%
        {wei2019twitter}
\bibfield{author}{\bibinfo{person}{Feng Wei} {and} \bibinfo{person}{Uyen~Trang
  Nguyen}.} \bibinfo{year}{2019}\natexlab{}.
\newblock \showarticletitle{Twitter bot detection using bidirectional long
  short-term memory neural networks and word embeddings}. In
  \bibinfo{booktitle}{\emph{TPS-ISA}}. IEEE, \bibinfo{pages}{101--109}.
\newblock


\bibitem[Yang et~al\mbox{.}(2019)]%
        {yang2019arming}
\bibfield{author}{\bibinfo{person}{Kai-Cheng Yang}, \bibinfo{person}{Onur
  Varol}, \bibinfo{person}{Clayton~A Davis}, \bibinfo{person}{Emilio Ferrara},
  \bibinfo{person}{Alessandro Flammini}, {and} \bibinfo{person}{Filippo
  Menczer}.} \bibinfo{year}{2019}\natexlab{}.
\newblock \showarticletitle{Arming the public with artificial intelligence to
  counter social bots}.
\newblock \bibinfo{journal}{\emph{Comput. Hum. Behav.}} \bibinfo{volume}{1},
  \bibinfo{number}{1} (\bibinfo{year}{2019}), \bibinfo{pages}{48--61}.
\newblock


\bibitem[Yang et~al\mbox{.}(2020)]%
        {yang2020scalable}
\bibfield{author}{\bibinfo{person}{Kai-Cheng Yang}, \bibinfo{person}{Onur
  Varol}, \bibinfo{person}{Pik-Mai Hui}, {and} \bibinfo{person}{Filippo
  Menczer}.} \bibinfo{year}{2020}\natexlab{}.
\newblock \showarticletitle{Scalable and generalizable social bot detection
  through data selection}. In \bibinfo{booktitle}{\emph{AAAI}},
  Vol.~\bibinfo{volume}{34}. \bibinfo{pages}{1096--1103}.
\newblock


\bibitem[Yang et~al\mbox{.}(2022)]%
        {yang2022rosgas}
\bibfield{author}{\bibinfo{person}{Yingguang Yang}, \bibinfo{person}{Renyu
  Yang}, \bibinfo{person}{Yangyang Li}, \bibinfo{person}{Kai Cui},
  \bibinfo{person}{Zhiqin Yang}, \bibinfo{person}{Yue Wang},
  \bibinfo{person}{Jie Xu}, {and} \bibinfo{person}{Haiyong Xie}.}
  \bibinfo{year}{2022}\natexlab{}.
\newblock \showarticletitle{RoSGAS: Adaptive Social Bot Detection with
  Reinforced Self-Supervised GNN Architecture Search}.
\newblock \bibinfo{journal}{\emph{ACM Transactions on the Web}}
  (\bibinfo{year}{2022}).
\newblock


\bibitem[Yardi et~al\mbox{.}(2010)]%
        {yardi2010detecting}
\bibfield{author}{\bibinfo{person}{Sarita Yardi}, \bibinfo{person}{Daniel
  Romero}, \bibinfo{person}{Grant Schoenebeck}, {et~al\mbox{.}}}
  \bibinfo{year}{2010}\natexlab{}.
\newblock \showarticletitle{Detecting spam in a twitter network}.
\newblock \bibinfo{journal}{\emph{First monday}} (\bibinfo{year}{2010}).
\newblock


\bibitem[Zhang et~al\mbox{.}(2022)]%
        {zhang2022fine}
\bibfield{author}{\bibinfo{person}{Lin Zhang}, \bibinfo{person}{Li Shen},
  \bibinfo{person}{Liang Ding}, \bibinfo{person}{Dacheng Tao}, {and}
  \bibinfo{person}{Ling-Yu Duan}.} \bibinfo{year}{2022}\natexlab{}.
\newblock \showarticletitle{Fine-tuning global model via data-free knowledge
  distillation for non-iid federated learning}. In
  \bibinfo{booktitle}{\emph{CVPR}}. \bibinfo{pages}{10174--10183}.
\newblock


\bibitem[Zhang(2022)]%
        {zhang2022feddtg}
\bibfield{author}{\bibinfo{person}{Zhenyuan Zhang}.}
  \bibinfo{year}{2022}\natexlab{}.
\newblock \showarticletitle{FedDTG: Federated Data-Free Knowledge Distillation
  via Three-Player Generative Adversarial Networks}.
\newblock \bibinfo{journal}{\emph{arXiv}} (\bibinfo{year}{2022}).
\newblock


\bibitem[Zhao et~al\mbox{.}(2020)]%
        {zhao2020multi}
\bibfield{author}{\bibinfo{person}{Jun Zhao}, \bibinfo{person}{Xudong Liu},
  \bibinfo{person}{Qiben Yan}, \bibinfo{person}{Bo Li},
  \bibinfo{person}{Minglai Shao}, {and} \bibinfo{person}{Hao Peng}.}
  \bibinfo{year}{2020}\natexlab{}.
\newblock \showarticletitle{Multi-attributed heterogeneous graph convolutional
  network for bot detection}.
\newblock \bibinfo{journal}{\emph{IS}}  \bibinfo{volume}{537}
  (\bibinfo{year}{2020}), \bibinfo{pages}{380--393}.
\newblock


\bibitem[Zhao et~al\mbox{.}(2018)]%
        {zhao2018federated}
\bibfield{author}{\bibinfo{person}{Yue Zhao}, \bibinfo{person}{Meng Li},
  \bibinfo{person}{Liangzhen Lai}, \bibinfo{person}{Naveen Suda},
  \bibinfo{person}{Damon Civin}, {and} \bibinfo{person}{Vikas Chandra}.}
  \bibinfo{year}{2018}\natexlab{}.
\newblock \showarticletitle{Federated learning with non-iid data}.
\newblock \bibinfo{journal}{\emph{arXiv}} (\bibinfo{year}{2018}).
\newblock


\bibitem[Zhu et~al\mbox{.}(2019)]%
        {DBLP:conf/emnlp/ZhuWWZZWZ19}
\bibfield{author}{\bibinfo{person}{Junnan Zhu}, \bibinfo{person}{Qian Wang},
  \bibinfo{person}{Yining Wang}, \bibinfo{person}{Yu Zhou},
  \bibinfo{person}{Jiajun Zhang}, \bibinfo{person}{Shaonan Wang}, {and}
  \bibinfo{person}{Chengqing Zong}.} \bibinfo{year}{2019}\natexlab{}.
\newblock \showarticletitle{{NCLS:} Neural Cross-Lingual Summarization}. In
  \bibinfo{booktitle}{\emph{EMNLP/IJCNLP}}. \bibinfo{publisher}{ACL},
  \bibinfo{pages}{3052--3062}.
\newblock


\bibitem[Zhu et~al\mbox{.}(2021)]%
        {zhu2021data}
\bibfield{author}{\bibinfo{person}{Zhuangdi Zhu}, \bibinfo{person}{Junyuan
  Hong}, {and} \bibinfo{person}{Jiayu Zhou}.} \bibinfo{year}{2021}\natexlab{}.
\newblock \showarticletitle{Data-free knowledge distillation for heterogeneous
  federated learning}. In \bibinfo{booktitle}{\emph{ICML}}. PMLR,
  \bibinfo{pages}{12878--12889}.
\newblock


\bibitem[Zia et~al\mbox{.}(2022)]%
        {zia2022improving}
\bibfield{author}{\bibinfo{person}{Haris~Bin Zia}, \bibinfo{person}{Ignacio
  Castro}, \bibinfo{person}{Arkaitz Zubiaga}, {and} \bibinfo{person}{Gareth
  Tyson}.} \bibinfo{year}{2022}\natexlab{}.
\newblock \showarticletitle{Improving Zero-Shot Cross-Lingual Hate Speech
  Detection with Pseudo-Label Fine-Tuning of Transformer Language Models}. In
  \bibinfo{booktitle}{\emph{ICWSM}}, Vol.~\bibinfo{volume}{16}.
  \bibinfo{pages}{1435--1439}.
\newblock


\end{thebibliography}

\appendix
\newpage 
\section{Appendix}
\label{sec:append}

\subsection{Glossary of Notations}
\label{sec:append:notation}
In Table~\ref{tab:notation}, we summarize the main notations used in this work.

\begin{table}[b]
\centering
\caption{Notations.}
\label{tab:notation}
\scalebox{0.8}{
\begin{tabular}{r|l}
\hline
\toprule
\textbf{Symbol}       & \textbf{Definition}  \\ 
\midrule
$\mathcal{L}$ & Defined loss function \\
$K$; $N_k$ & Number of clients; Number of samples in $k$-th client \\
$x_m$; $y_n$ & The $m$-th word and $n$-th word in two different language contents  \\
$\phi_E$ & The encoder for text content representation \\
$\phi_D$ & The decoder for translating context representation \\
$\mathcal{M}$ & The cross-lingual mapper for converting context representation \\
$z_x$  & Text context representation vector \\
$\epsilon$ & The backbone model for user feature extraction \\
$u_t$ & The tweets set from $t$-th user \\ 
$t^m_i$ & The $i$-th word in user's $m$-th tweet \\
$h^t_m$ & Representation for $m$-th tweet posted by $t$-th user   \\
$r_p$; $r_t$ & User's property representation; tweet-level representation \\
$r_u$,$r$ & User-level representation \\
$G$; $D_1$, $D_2$ & Generator; Discriminators \\
$\mathcal{D}_k$ & The local data stored in the $k$-th client \\
$\mathcal{D}$ & The data set collected from all clients  \\
$(x^k_i,y^k_i)$ & The $i$-th sample pair stored in the $k$-th client \\
$\mathcal{N}(0,1)$ & The gaussian distribution \\
$\tilde{x}$ & The pseudo-data generated by generators \\
$p$; $\hat{p}$ & Probability of local data; Global data distribution probability \\
$D_{KL}$ & The Kullback–Leibler divergence \\
$\tau$ & The temperature parameter for smoothing similarity value \\
$\gamma,\mu$ & The weight hyperparameter in the defined loss function \\
$\theta$ & The parameters of the designed model \\
$M$ & The number of clients participating in the communication \\
$\alpha^{k,y}_t$ & Ratio of samples with label $y$ stored in ${client}_k$ against in $\mathcal{D}$ \\

\bottomrule
\end{tabular}}
\end{table}

\subsection{Statistics of Datasets}
\label{sec:append:datasets}

Table~\ref{tab:dataset-statistic} summarizes the
basic statistic information of datasets used in this work.

\begin{table}[b]
\setlength{\abovecaptionskip}{0.15cm}
\setlength{\belowcaptionskip}{-0.35cm}
\caption{Statistics of datasets.}
\vspace{-0.5em}
\label{tab:dataset-statistic}
\centering
\scalebox{0.9}{
\begin{tabular}{c|c|c|c}
\hline
\toprule
\specialrule{0em}{1pt}{1pt}
Dataset   & Humans   & Bots    & Total number   \\ 
\specialrule{0em}{1pt}{1pt}
\hline
\specialrule{0em}{1pt}{1pt}
Vendor-19 \cite{yang2019arming} & 1860 & 568  & 2428 \\
TwiBot-20 \cite{feng2021twibot} & 4175 & 5286 & 9461 \\
\specialrule{0em}{1pt}{1pt}
\bottomrule
\hline
\end{tabular}
}
\end{table}

\subsection{Generalization Experiment}
\label{sec:append:a}

To validate the generalization of our method, we collected four additional public social bot detection datasets: Varol-17, Gilani-17, cresci-19, botometer-feedback-19. To simulate the scenario of bot detection in uniting multiple social platforms, we select one dataset from Vendor-19 and TwiBot-20 as the test dataset. Each of the remaining datasets is distributed to a specific client. In this circumstance, there are five clients and a server, and each client represents a social platform. 
We aim to demonstrate the enhanced model proposed in this work can still competitively detect new variants that have never been seen before, when all platforms share the characteristics of their own social bots with other platforms. 
As shown in Table~\ref{tab:test-generalization-table}, the difficulty detection in the two datasets is entirely different. For most of the approaches, the classification accuracy of TwiBot-20 dataset is merely 50\%, which indicates a huge discrepancy of bot features among datasets. In other words, the detection models learnt from the early generations of bots can hardly detect the bots in TwiBot-20 in an accurate manner. By contrast, our approach can achieve a higher accuracy due to the ability to learn a consistent feature space; as a result, the bot account features in different clients can be effectively shared. Additionally, constraining the optimization direction of the client model can facilitate our model to obtain better representation ability, and hence a better generalization  than others.

\begin{table}
\setlength{\tabcolsep}{6mm}{
\caption{Comparison of the average maximum accuracy of different methods which are tested on a specific dataset (Vendor-19 or TwiBot-20) and trained from the other datasets.}
\label{tab:test-generalization-table}
\centering
\scalebox{1}{
\begin{tabular}{@{}c|c|c@{}}
\hline
\toprule
Test Dataset             &   Vendor-19             &     TwiBot-20 \\ 
\specialrule{0em}{1pt}{1pt}
\hline
\specialrule{0em}{1pt}{1pt}
FedAvg              &  76.53$\pm$0.11   & 49.63$\pm$1.09 \\
FedProx             & 74.26$\pm$3.57      & 54.21$\pm$2.01  \\
Ensemble            & 76.01$\pm$0.65       & 51.98$\pm$5.50   \\
FedDistill          & 73.11$\pm$0.95      & 51.19$\pm$1.12    \\
FedGen              & 77.52$\pm$0.65       & 62.30$\pm$0.93   \\
FedFTG              & 76.51$\pm$1.15     & 60.12$\pm$1.01 \\  
\specialrule{0em}{1pt}{1pt}
\hline
\specialrule{0em}{1pt}{1pt}
\FedACK         & \textbf{78.13$\pm$1.38}      & \textbf{64.79$\pm$1.61}   \\
\bottomrule
\hline
\end{tabular}}
}


\end{table}

\end{document}